\pdfoutput=1
\documentclass[11pt]{article}

\usepackage{xcolor}

\usepackage[final]{acl}

\usepackage{times}
\usepackage{tabularx}
\usepackage{booktabs}
\usepackage{cuted}
\usepackage{longtable}
\usepackage{supertabular}
\usepackage{makecell}
\usepackage{array}
\usepackage{graphicx}
\usepackage{adjustbox}
\usepackage{hyperref}

\usepackage{listings}
\lstdefinestyle{pythonstyle}{
    language=Python,
    basicstyle=\ttfamily\small,
    breaklines=true,
    numbers=left,
    numberstyle=\tiny\color{gray},
    frame=single,
    rulecolor=\color{black},
    keywordstyle=\color{blue}\bfseries,
    commentstyle=\color{green!60!black}\itshape,
    stringstyle=\color{red},
}

\lstdefinestyle{textstyle}{
    basicstyle=\ttfamily\footnotesize,
    breaklines=true,
    frame=single,
    rulecolor=\color{black},
}

\usepackage{siunitx}
\usepackage{latexsym}
\usepackage{amsmath}
\usepackage{yhmath}
\usepackage{dsfont}
\usepackage{float}
\usepackage[T1]{fontenc}
\usepackage[utf8]{inputenc}

\usepackage{microtype}

\usepackage{inconsolata}

\usepackage{graphicx}
\usepackage{xspace}

\usepackage{color-edits}
\addauthor{gn}{magenta}
\addauthor{ab}{teal}
\addauthor{ar}{blue}

\newcommand{\nummodels}{92\xspace}

\title{Not-Just-Scaling Laws: Towards a Better Understanding of the Downstream Impact of Language Model Design Decisions}

\newcommand{\medsize}{\fontsize{9.5pt}{10pt}\selectfont}

\author{
  \textbf{Emmy Liu\textsuperscript{1}},
  \textbf{Amanda Bertsch\textsuperscript{1}},
  \textbf{Lintang Sutawika\textsuperscript{1}},
  \textbf{Lindia Tjuatja\textsuperscript{1}},
  \textbf{Patrick Fernandes\textsuperscript{1,2,3}}\\
  \textbf{Lara Marinov\textsuperscript{1}},
  \textbf{Michael Chen\textsuperscript{1}},
  \textbf{Shreya Singhal\textsuperscript{1}},
  \textbf{Carolin Lawrence\textsuperscript{4}}, \\
  \textbf{Aditi Raghunathan\textsuperscript{1}},
  \textbf{Kiril Gashteovski\textsuperscript{4, 5}},
  \textbf{Graham Neubig\textsuperscript{1}}\\
  \medsize{\textsuperscript{1} Carnegie Mellon University, Language Technologies Institute} \vspace{-2pt}\\
  \medsize{\textsuperscript{2}Instituto Superior Técnico  (Lisbon ELLIS Unit), \textsuperscript{3} Instituto de Telecomunicações} \vspace{-2pt}\\
  \medsize{\textsuperscript{4} NEC Laboratories Europe, Germany} \vspace{-2pt}\\
  \medsize{\textsuperscript{5} CAIR, Ss. Cyril and Methodius University of Skopje, North Macedonia} \vspace{-2pt}\\
  \small{\textbf{Correspondence:} \href{mailto:emmy@cmu.edu}{emmy@cmu.edu}}
}

\begin{document}
\maketitle
\begin{abstract}

%\gncomment{For the title, one other idea: ``Not-just-scaling Laws: Towards Finer-grained Understanding of the effect of Language Modeling Design Decisions on Downstream Performance''. This is more wordy, but might be a little bit more compelling.} \abcomment{what about ``Not-just-scaling Laws: Towards finer-grained understanding of the downstream impact of language modeling design decisions''?}

%\arcomment{I think ``scale'' is a bit ambiguous, and we should define it at least once concretely. Could be nice to give examples of scale and non-scale features}
Improvements in language model capabilities are often attributed to increasing model size or training data, but in some cases smaller models trained on curated data or with different architectural decisions can outperform larger ones trained on more tokens. What accounts for this?
To quantify the impact of these design choices, we meta-analyze \nummodels open-source pretrained models across a wide array of scales, including state-of-the-art open-weights models as well as less performant models and those with less conventional design decisions. We find that by incorporating features besides model size and number of training tokens, we can achieve a relative 3-28\% increase in ability to predict downstream performance compared with using scale alone.
Analysis of model design decisions reveal insights into data composition, such as the trade-off between language and code tasks at 15-25\% code, as well as the negative association of web data with truthfulness.
Broadly, our framework lays a foundation for more systematic investigation of how model development choices shape final capabilities.
\footnote{Code and data are available at \url{https://github.com/nightingal3/llm-pretraining-behaviours/}}
%\footnote{Code and data are available at \url{https://github.com/nightingal3/llm-pretraining-behaviours} for the community to build upon.}

%\gncomment{we can come up with a more interesting initial sentence than this}.
\end{abstract}

\section{Introduction}
%\arcomment{Be careful to substantiate what exactly scaling laws mean. People do design scaling laws for data-mixtures and try to extrapolate what's the right combination of data to use for e.g., so how would that fit in here?} 

The effectiveness of language model training depends critically on decisions made during pretraining. For instance, the effectiveness of scaling up data depends on its composition -- processing even a trillion tokens would be ineffective if they all consisted of the word ``the''. Language model performance has been found to be fairly predictable through \emph{scaling laws} (\citet{kaplan2020scaling}, \autoref{sec:scaling_laws}) -- extrapolations of model performance based on the parameter counts and number of tokens the models were trained on. However, scaling laws based on only these two aspects do not always explain downstream task performance \cite{diaz2024scalinglawsscale, isik2024scaling}.
%\arcomment{This sounds like the main factor is the gap between pretraining perplexity and downstream performance. Is is true that additional features help address this gap, or just help predict pretraining perplexity better too?}
\begin{figure}[!ht]
    \centering
    \includegraphics[width=0.75\linewidth]{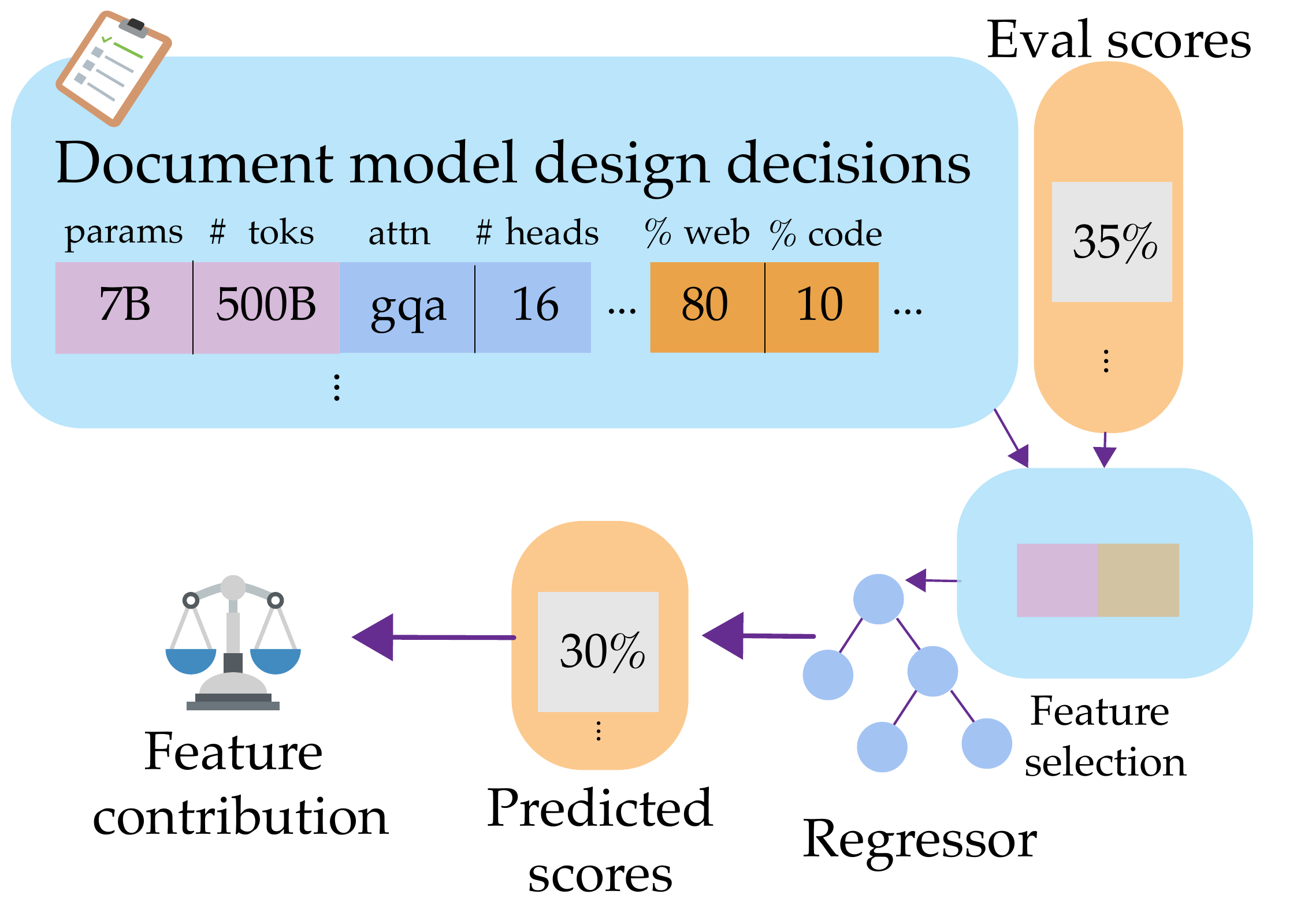}
    \caption{We document design decisions from open-weights models related to both architecture and data composition, and train predictors for downstream task performance. This allows us to examine the impact of model design choices individually.\vspace{-1.5em}}
    \label{fig:method-illustration}
\end{figure}
The research community has made progress in understanding how training decisions impact downstream performance with respect to data composition. For instance, controlled studies have demonstrated that training on code data improves performance on certain reasoning benchmarks \cite{aryabumi2024codecodeexploringimpact, petty2024doescodepretrainingaffect}, meta-features of data such as age and the use of toxicity filters affect performance on many QA tasks \cite{longpre2023pretrainer}, and the balance of multilingual data affects performance on English and other languages \cite{chang2023multilingualitycurselanguagemodeling, pangea}.
These works uncover valuable insights, but they tend to focus on changing only a single aspect of the training recipe while keeping the rest fixed. Although rigorous, this is costly in compute and development time.
% All else being equal, when training a model with fixed characteristics, we would like to ideally estimate how much data to train a model on, but also what that model's expected final performance is on downstream tasks without needing to train the model. Additionally, while some data may be higher quality in general and more broadly useful, there are also always tradeoffs to be made when selecting data broadly -- one obvious one is that when optimizing for English language benchmarks, data optimization methods will naturally select English data, as will model developers when intuitively choosing pretraining data. This will obviously hurt multilingual performance, in much the same way as choosing too much multilingual data may degrade performance on English benchmarks. This tradeoff may be extremely obvious, but there may be less obvious tradeoffs as well within a fixed data budget, as the inclusion of one document means that we cannot include another, leading to a model with different characteristics and hence different downstream performance, even for data commonly seen as high quality.
%\arcomment{What is holistic? Do you mean you actually model joint interactions? if so, we can state more clearly} 
%In this paper we instead take a \emph{holistic} approach, by pooling past findings from open language models across multidimensional design decisions, and examining how these decisions jointly affect downstream performance.  
We instead ask: can we leverage past findings from open language models to examine which training decisions are associated with better downstream performance?
%\arcomment{I'd REALLY play this up here.. Maybe we can even state it more directly: people train their own scaling laws specialized to their setups. They can only feasibly vary a couple parameters. But there is a trove of information out there already - what can we learn from all these models that have been trained? There MUST be something interesting we can discover from this collective experiments}

To do so, we first \emph{catalog} features regarding the model architecture, and data of \nummodels base pretrained LMs from varied families ($\S\ref{sec:database}$).
The resulting database of model features spans most major original pretrained decoder-only models released open-weights between the years 2019-2024.

We then develop methodology to \emph{predict performance} of these models across a wide array of benchmarks both based on traditional scaling factors as well as architectural decisions and data composition ($\S\ref{sec:prediction}$).
Specifically, we train regression models that take in the extracted features and predict the benchmark results, and further use model interpretability techniques to identify the most salient features in making these predictions.

We evaluate this methodology on predicting performance across 12 popular LLM benchmarks, and demonstrate that it is \emph{not just scaling} that determines model performance -- on all benchmarks the regressor with all features outperforms a regressor based solely on scaling model features ($\S\ref{sec:predictor_performance}$).
Our analysis of feature importance reveals potential impacts of data domains on task performance, 
reconfirming empirical results such as the best ratio of code to use in pretraining ($\S\ref{sec:what_features}$). Furthermore, we find that features extracted from a model’s generated text -- such as the frequency of question-related words or the proportion of web-like text—help predict performance on various benchmarks. This suggests that a model’s generation patterns can reflect underlying biases from its pretraining data that, in turn, influence downstream performance.

%\arcomment{Are you describing all the insights here?} \arcomment{Did not understand the next sentence even after reading twice. This isn't something we can tell before pretraining a model right?} 
%\arcomment{A bit weak.. I'd focus on the data collection we are putting out and say people can do more analysis on this and we hope this is a valuable resource.} 

By documenting open-source models trained by the entire community and extracting insights, we provide a practical resource for model developers to learn from collective experience. We discuss this and future work in $(\S\ref{sec:conclusion})$.

\section{Scaling Laws}

% \newcommand\figheight{2in}
% \begin{figure}[t]
%   \centering
%     \begin{minipage}[c][\figheight][c]{1.0\textwidth}
%     \centering
%     \includegraphics[height=1.8in,width=\textwidth]{figures/scaling_eyecatcher.pdf}
      
%     \caption{When considering pretrained base model performance, different tasks can have different scaling factors, and individual model families or models can also deviate from the overall pattern due to individual factors.}%gncomment{I like this figure a lot, but I'm a little bit worried about whether it'll be easy enough for people to understand to put it on the first page. Alternatively we could have a more ``methods'' style figure where we just demonstrate what our method is.}
%     \label{fig:eyecatcher-scaling}}
%     \end{minipage}%
% \end{figure}
% % \begin{figure}[t]\vspace{\figheight}\end{figure}

\label{sec:scaling_laws}

\subsection{Definition}

We define scaling laws here as a relationship between the number of parameters $N$ and the number of tokens $D$ of a language model family, and the expected language-modelling loss at convergence $L(N, D)$.\footnote{Please see $\S\ref{sec:rw-other-scaling-laws}$ for more detailed discussion; scaling laws can and do take into account other factors in various works, but for simplicity we call $N$ and $D$ scale-related here, while all other decisions in $\S\ref{sec:features}$ are contrasted with these.} 
Importantly, these laws are typically examined while holding all other factors constant: keeping the same model architecture, training data, and model parameters. 
Originally, \citet{kaplan2020scaling} showed that over a wide range of transformer-based models, this relationship can be expressed as a power law:

\begin{equation}
\small
\label{eq:kaplan-scaling}
    L(N, D) = \left( \left(\frac{N_c}{N}\right)^{\frac{\alpha_N}{\alpha_D}} + \frac{D_c}{D} \right)^{\alpha_D}
\end{equation}
Later, \citet{chinchilla} introduced a similar law, which differed in the coefficients fitted, but was also based on a power law.

However, scaling laws are not absolute, and the exact functional form and fitted coefficients may depend on the architecture type, size range \cite{pearce2024reconcilingkaplanchinchillascaling}, or other considerations such as inference costs. See ($\S\ref{sec:rw-other-scaling-laws}$) for further discussion. 
%\gncomment{I wonder how important th information in this paragraph is to the story? I feel like it could possibly be deleted.}

%%%
% \begin{itemize}
%     \item{Introduce basic Kaplan/Chinchilla scaling laws}
%     \item{Assumptions made/things they don't account for}
% \end{itemize}

\subsection{Maybe it's Not Just Scaling?}

%\gncomment{I did a bit of writing here, please take a look.}

Are parameter count and number of training tokens really all that are needed to accurately predict a model's downstream performance?
Intuitively the answer is ``no'' -- there are a number of design decisions that go into model training, and all of them could have an effect on model performance. 

\paragraph{Model Architecture Details}
While the majority of modern language models follow the transformer architecture, there are some details that differ.
For instance, the variety \citep{rmsnorm} and position \citep{postnorm_vs_prenorm} of layer normalization, and the type of positional encoding \citep{rope,alibi} make significant differences in model performance.
Previous work, such as \citet{mamba}, has demonstrated empirically that holding all other factors equal, models that make better architecture decisions \citep{touvron2023llama} outperform those that make worse decisions \citep{vaswanitransformer}.

\paragraph{Data Composition}
%Scaling laws can also be made to account for a range of other considerations, such as inference costs over a model's lifetime \cite{},  quantization \cite{}, data quality \cite{}, and data composition \cite{}. In particular, data quality plays a crucial role. 
% Although typical scaling laws are fitted on large and diverse pretraining datasets, which may include web data, books, source code and more, varying the composition and quality of where tokens come from inevitably impacts the final model. For instance, even if a model is trained to convergence with an ``optimal'' number of tokens, its performance may deviate from the scaling law if the token distribution differs substantially from that used to fit the law. As an extreme (humorous) example, training on 1T tokens of just the word ``the'' is unlikely to yield good performance.
In addition, data composition and quality plays a major role in the final quality of a model.
For instance, past work has demonstrated that training on some quantity of code improves performance on English reasoning tasks \citep{ma2023trainingstagedoescode}.
Also, work has demonstrated that filtering for ``educational'' content allows for more efficient learning and higher performance on knowledge-based question answering tasks \citep{textbooks_are_all_you_need}.

%\paragraph{Training Algorithm}
%There are also a number of factors related to training, such as the optimization algorithm and learning rate, that affect final model performance.

\paragraph{Task Setting}
Finally, there is an interplay of all the above factors with how model performance is measured.
While previous work on scaling laws has mostly measured loss values, downstream users usually care about task performance, rather than validation loss on a pretraining dataset.
Although there is often a correlation between the two for many tasks, certain tasks may be harder to predict from a model's loss alone \cite{bhagia2024establishingtaskscalinglaws}. Moreover, certain tasks exhibit pathological scaling behaviour, such as inverse or U-shaped scaling \cite{caballero2023brokenneuralscalinglaws, wei2023inversescalingushaped, mckenzie2024inversescalingbiggerisnt}, or simply more unpredictable performance \cite{isik2024scaling}. 
%This may be because performance on these tasks is dependent on other factors such as data quality, phase transitions in representations, or <smth else> thus producing an odd trend when slicing performance along size-based dimensions.

We ask: \emph{can we more effectively predict the performance of LLMs by devising a new set of ``laws'' that are not just reliant on scaling-based factors?}

% examine scaling performance across a diverse set of tasks, in order to understand the differences in scaling behaviour across tasks and also choices made by individual models that lead them to deviate from their expected performance. 
%%%
% \begin{itemize}
%     \item{Variations on power-law functional form (e.g. adding terms for domain composition)}
%     \item{Other variations trying to account for different things, e.g. finetuning or quantization (maybe move to related work)}
% \end{itemize}

\section{Building a Database of Publicly-Available Language Models}
\label{sec:database}

To address our research question, we built a database of publicly available language models spanning 11M to 110B parameters,\footnote{Including embedding parameters.} limited to distinct pretrained decoder-only base models.\footnote{By distinct, we mean unique combinations of training data and architecture. Models trained on deduplicated datasets are counted separately, but not variants with different curricula/initializations.} This section describes our inclusion criteria, model featurization, and evaluation approach.

\subsection{Data Collection}

To ensure that our analysis was consistent, we applied the following criteria:

\paragraph{Pretrained-only:} Only base models that were pretrained from scratch were included. Fine-tuned variants, merged models, and models with additional post-training were excluded. 

\paragraph{Architecture:} Only transformer-based decoder-only models were included to maintain uniformity. Mixture-of-experts (MoEs) or other architectures were excluded.

\paragraph{Publicly available information:} Only models with publicly available metadata, documented through configuration files or papers, were included. In particular, both the total number of parameters and total number of tokens trained on were required for inclusion. A full list of models and model families can be found in \autoref{appendix:all_models}.

%Models included range from 10M to 100B parameters. 

% \begin{itemize}
%     \item{How we filtered models for inclusion (pretrained base models, no merges, no MoEs, decoder-only architecture)}
%     \item{Featurizing model design decisions: model architecture vs pretraining data design}
%     \item{Evaluation suite and how it was chosen (mostly due to open-llm-leaderboard having lots of evals already done, but also a reasonable diversity)}
% \end{itemize}

\subsection{Characterizing Models and Data}
\label{sec:features}

\begin{figure}[t]
    \centering
    \includegraphics[width=0.9\linewidth]{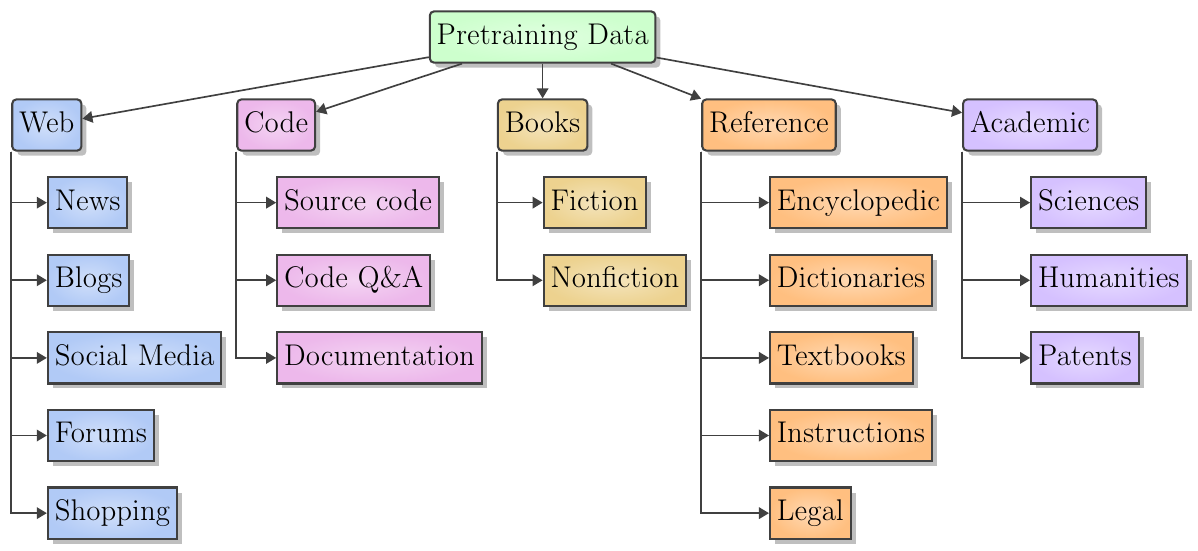}
    \caption{Taxonomy of pretraining data categories. We sorted data sources into this taxonomy based on model documentation.}
    \label{fig:data-taxonomy}
\end{figure}

We represent each model by the architectural choices it makes, as well as its choice of pretraining data. Formally, let $\mathcal{A}$ be the set of features related to model architecture, and $\mathcal{D}$ be the set of features related to the model's pretraining dataset. For each task $T$ we want to approximate a model $M$'s true score $s_T$ with a prediction $\widehat{s_T}$:
\begin{equation}
    \small
    \widehat{s_T}(M) = f_{\theta}([\mathcal{A}_M;\mathcal{D}_M]).
\end{equation}
This reduces to typical scaling laws when $\mathcal{A} = \{ \text{\# params} \} $, $\mathcal{D} = \{\text{\# tokens}\}$, and $f_{\theta}$ is a power law.

In total, we document \nummodels open models along the dimensions of model features, high-level dataset features, and features derived from that model's no-context generations.
For the full set of features and definitions, please see \autoref{appendix:all_features}.

\subsubsection{Features from Model Documentation}

We first collect information about each model by reading source papers/blogs when available (see \autoref{appendix:all_models} for original citations), as well as data listed on the Hugging Face Hub \cite{transformers-lib}.

%\el{TODO: move this into 4.1 and 4.2 resp.}
%We start by examining the architectural decisions and dataset features used in pretraining open-source language models ranging from 125M to 175B parameters, focusing on models before any post-training procedures such as instruction tuning or preference tuning. We collect the following information about each model by reading source papers when available, and falling back to data listed on the huggingface hub if no paper was available for that model. In total, we document 126 open models along the dimensions of model features, high-level dataset features, and document-level dataset features, focusing on models with a decoder-only transformer-based architecture. Although there are some prominent open-source models based on alternative architectures (most commonly state-space models or RNN-like models), these are comparatively less common, and may have a disjoint set of model features that need to be documented.
\paragraph{Architectural Features:}
These features capture design decisions that determine model structure. For example, \textit{total parameters} (including embedding parameters), the number of transformer layers, the embedding and feed-forward dimensions, and details such as the type of layer normalization or attention variant used.

\paragraph{Data Features:}
These features summarize pretraining data composition. Representative examples include \textit{total tokens trained on} and the percentage breakdown of tokens sourced from various domains defined in \autoref{fig:data-taxonomy}, as well as the proportion of English-language tokens. Our pretraining data domains were derived from common subdomains in open pretraining datasets \cite{pile, dolma}. We use the top level domains (web, code, books, reference, academic) as this tends to be the granularity at which data composition is described in papers.

\subsubsection{Exploring Data Composition via Generation}
Although many models document some data composition details, relatively few release their full pretraining corpus, resulting in missing values for many models in our study.

We propose estimating a model's training data by generating from it with only a beginning-of-sequence token in context (or end-of-sequence if BOS is unavailable). Using temperature-based sampling with $T = 1$,\footnote{this should in principle recover $P_{raw}(x_t \mid x_{<t}) \approx P(x_t \mid x_{<t})$ in the limit of sampling infinitely from an LM that captures its distribution perfectly.} we sample 5-10k free-generations per model and categorize them using an LM-based classifier (see \autoref{appendix:freegen-prompt}) according to the domains in \autoref{fig:data-taxonomy}. A human annotator independently validated the classifier on 300 documents from pretraining datasets (\autoref{appendix:classifier_validation}), showing high agreement (Cohen's $\kappa=0.746$).

We also extract lower-level linguistic features such as words per sentence, constituency tree depth, and dependency length. Our validation analysis (\autoref{appendix:freegen_validation}) shows that domain-level features correlate well with actual pretraining data composition (e.g., web content correlation: $r=0.916$, $p=7.55\times10^{-12}$), while lower-level stylistic features show weaker correlations. However, holistic model-level correlations across all features are strong (typically $r>0.8$), supporting our use of free-generations as proxies for pretraining composition, while not substituting free-generation features for pretraining features.

\subsection{Evaluation Datasets and Metrics}
\begin{table}[t]
\centering
\renewcommand{\arraystretch}{0.9} % reduce row spacing
\resizebox{0.7\columnwidth}{!}{
\begin{tabular}{l l l}
\toprule
\multicolumn{3}{c}{\textbf{Commonsense Reasoning / NLI}} \\
\midrule
ANLI \cite{nie2020adversarial}         & $\sim163$k     & Brier Score \\[0.2em]
HellaSwag \cite{zellers2019hellaswag}    & $\sim70$k      & Accuracy \\[0.2em]
Winogrande \cite{sakaguchi2019winogrande} & $\sim44$k      & Accuracy \\[0.2em]
XNLI \cite{conneau2018xnli}              & $\sim2.5$k     & Brier Score \\
\midrule
\multicolumn{3}{c}{\textbf{Math / Logic}} \\
\midrule
GSM8K \cite{cobbe2021gsm8k}        & $8\,000$       & Accuracy \\[0.2em]
LogiQA2 \cite{wang2020logiqa}      & $\sim8$k       & Brier Score \\[0.2em]
MathQA \cite{saxton2019mathqa}      & $\sim37$k      & Brier Score \\
\midrule
\multicolumn{3}{c}{\textbf{General Knowledge}} \\
\midrule
ARC Challenge \cite{clark2018arc} & $\sim2.6$k     & Accuracy \\[0.2em]
Lambada \cite{paperno2016lambada}   & $\sim10$k      & Accuracy \\[0.2em]
MMLU \cite{hendrycks2020measuring}  & $\sim2.85$k    & Accuracy \\
\midrule
\multicolumn{3}{c}{\textbf{Other}} \\
\midrule
TruthfulQA \cite{lin2021truthfulqa}   & $817$         & Accuracy \\[0.2em]
HumanEval \cite{chen2021evaluating}   & $164$         & Accuracy \\
\bottomrule
\end{tabular}
}
\caption{Overview of LM evaluation datasets with approximate sample counts and evaluation metrics. Datasets ANLI, XNLI, LogiQA2, and MathQA use Brier Score, while the others use Accuracy.\vspace{-1.9em}}
\label{tab:datasets}
\end{table}

To assess how design choices affect reasoning capabilities, we evaluated models on datasets from the Open LLM leaderboard \cite{openllm} that capture diverse aspects of reasoning (\autoref{tab:datasets}).\footnote{We excluded Arithmetic and Minerva math tasks \cite{brown2020language, minervamath} as we focused on base models, and few achieved non-zero scores.}
We collect results for some models directly from the leaderboard, and for models not on the leaderboard we use the Eleuther LM eval harness \citep{eval-harness} to conduct evaluations with exactly the same setting. In addition, if there were multiple versions of a task or sub-tasks, we evaluated all of them and averaged them to get the overall task score.
For the full list of evaluation datasets and settings, see \autoref{appendix:all_evals}.
% Our evaluation focuses on two key metrics:
%, namely \textbf{factual and common-sense reasoning} (MMLU, TruthfulQA, Winogrande), \textbf{mathematical reasoning} (GSM8k), and \textbf{general natural language comprehension and inference} (Lambada). 

For an evaluation dataset $T$ where the $i$-th sample is $y_i$ and model $M$, we define $s_T(M)$ with:

\paragraph{Accuracy}
We use unnormalized, exact-match accuracy $s_{T,\text{acc}} = \frac{1}{|T|} \sum_{i = 1}^{|T|} \mathds{1}{\{y_i = \hat{y_i}\}}$ for the majority of tasks. We use pass@1 for Humaneval, but group it with accuracy tasks for convenience.
% We remove tasks that show evidence of emergent behaviour from the accuracy score.

\paragraph{Brier score}
For tasks where smaller models struggle to achieve non-zero accuracy, we follow \citet{emergencemirage} in using multiclass brier score as an alternate continuous metric for multiple-choice tasks \cite{brier}.\footnote{Note that lower is better for brier score. Multiclass brier score ranges between 0-2.} For a task with $K$ classes, let $p_{ik}$ be the predicted probability for class $k$ on sample $i$. Then $s_{T, BS} = \frac{1}{|T|} \sum_{i = 1}^{|T|} \sum_{k = 1}^{K} (p_{ik} - \mathds{1}{\{y_i = k\}})^2$. 
%\abcomment{cite brier score?}

%\subsubsection{Held-out models}

\begin{figure*}
    \centering
    \includegraphics[width=0.8\textwidth]{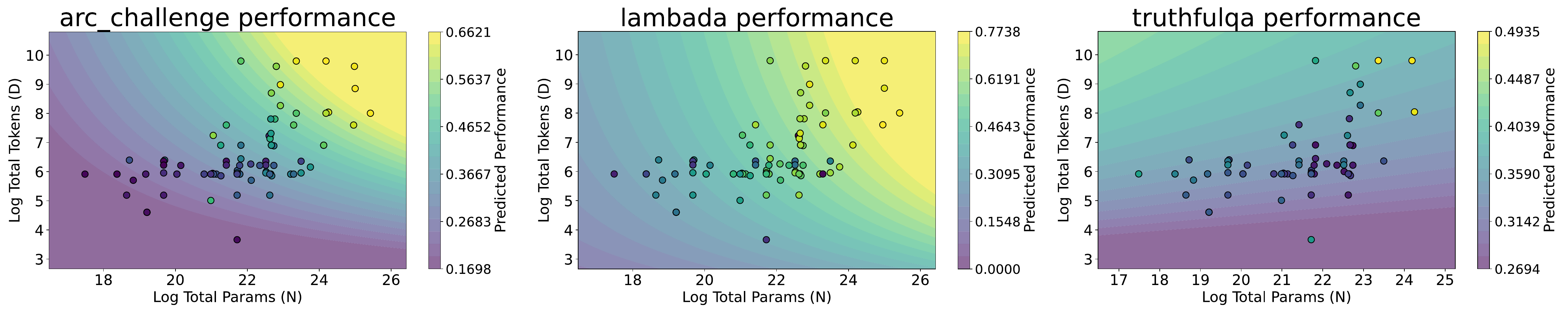}
    \caption{Performance of plotted against their total parameters and tokens. The background colour represents \autoref{eq:kaplan-scaling} fitted to the task, and the marker colours indicate true performance. Some tasks have different performance trends with scale. Within each task, individual models may also perform unexpectedly.}%gncomment{I like this figure a lot, but I'm a little bit worried about whether it'll be easy enough for people to understand to put it on the first page. Alternatively we could have a more ``methods'' style figure where we just demonstrate what our method is.}
    \vspace{-1em}
    \label{fig:eyecatcher-scaling}
\end{figure*}

\subsection{Heterogeneity in Task-specific Scaling}

%\gncomment{This discussion seems out of place here, we should probably put it in the results section.}

Before adding in other factors, we examine differences in scaling along $N$ and $D$ between our selected tasks. We fit a \citet{kaplan2020scaling} style law to each task. As seen in \autoref{fig:eyecatcher-scaling}, we see that different tasks may exhibit marked differences both in how well they follow scaling trends, as well as their individual scaling contours. For instance, TruthfulQA appears to exhibit U-shaped scaling, while Humaneval has more ``outlier'' models. A full list of $R^2$ values for tasks can be found in \autoref{appendix:task-deviation}.

\section{Predictive Modeling}
\label{sec:prediction}

%namely linear regressors based on the log of the number of parameters and training tokens \gncomment{need to double-check that I'm not lying here}.
Next, given our database we fit a regressor to try to predict performance.
In traditional scaling laws, regressors are fit based on power laws.
However, we are now dealing with a larger number of features, some of which may not be captured well by simple parametric forms.
Hence, we follow previous work on performance prediction \citep{xia-etal-2020-predicting, ye-etal-2021-towards} utilizing tree-based regressors based on XGBoost \cite{xgboost}.%
\footnote{We also performed preliminary experiments with LightGBM \cite{lightgbm} but it yielded very similar results in both prediction accuracy and feature importance. The LightGBM version of the main results can be found in \autoref{appendix:lightgbm_ver}.}

For each evaluation benchmark, we train a model to predict the performance metric on that task based on architectural features $\mathcal{A}$ and data features $\mathcal{D}$.
For each task setting, we perform 3-fold cross-validation due to the relatively small number of models,  with a nested inner cross-validation over the training set in each fold. The inner cross-validation conducted grid search over a small set of hyperparameters, allowing the model to slightly vary per task. See \autoref{appendix:xgboost_settings} for more details.

\paragraph{Evaluation}
To evaluate the predictors, we use Mean Absolute Error averaged across all models and folds. In other words, for a task with $N$ models evaluated, $\text{MAE}_T = \frac{1}{|T|} \sum_{i = 1}^{N} |s_T(M_i) - \widehat{s_T}(M_i)|$. We compare the scaling-laws predictor as well as the all-features predictor against each other, but also against the \textbf{median baseline}, which simply predicts the median score of the models in the training set for each model in the test set of that fold, and the \textbf{log-linear baseline}, which fits a log-linear function to the number of parameters and number of tokens.

\paragraph{Iterative Feature Selection}
As the full set of features is very large, we sequentially selected features from the full set greedily based on which reduced MAE the most, averaged across 5 random seeds.
Features were added until no reduction of at least $1\times10^{-4}$ was observed. 
We started using only the two scaling laws features, and refer to this as the \textbf{scaling-laws} model, though it does not have the form of a traditional power law.%
\footnote{As we use a tree-based predictor to accommodate diverse feature types (including non-numeric ones), our approach prioritizes interpolation within observed bounds (10M-100B parameters, 50B-3T tokens) rather than extrapolation. Exploring other prediction methods remains future work.}
By then incorporating additional architectural or data features, we can then directly quantify the incremental predictive power afforded by these extra features.
We refer to the model with the set of features as the \textbf{all-features} model. In all cases, we ran models with the same hyperparameter grid and the same random seeds and splits.

% , while noting that we can also experiment with ablating any features in $\mathcal{A}$, $\mathcal{D}$, or dropping one of these feature sets altogether.

\paragraph{Significance Testing}
Because the relative difference between baselines is small, we test both predictors across many seeds (50). We then ran paired t-tests on the overall MAE values for each seed, and corrected for multiple comparisons across tasks with the False Discovery Rate \cite{fdr}.

% Macro for printing MAE in percentages with ± CI on the same line.
\newcommand{\pmae}[2]{#1\% {\scriptsize{$\pm$ #2\%}}}
% Bold version (for lower MAE); add a star if significant.
\newcommand{\pmaeB}[2]{\textbf{#1\% {\scriptsize{$\pm$ #2\%}}}}
\newcommand{\pmaeBS}[2]{\textbf{#1\% {\scriptsize{$\pm$ #2\%$^\ast$}}}}

\newcommand{\mae}[2]{#1 {\scriptsize{$\pm$ #2\%}}}
\newcommand{\maeB}[2]{\textbf{#1{\scriptsize{$\pm$ #2\%}}}}
\newcommand{\maeBS}[2]{\textbf{#1{\scriptsize{$\pm$ #2\%$^\ast$}}}}

% Macro for formatting p-values using siunitx.
\newcommand{\pval}[1]{\num[exponent-product=\times]{#1}}

\begin{table*}[!ht]
\centering
\small
\resizebox{0.9\textwidth}{!}{
\begin{tabular}{ll c c c c c}
\toprule
\textbf{Benchmark} & \textbf{Setting} & \textbf{Baseline MAE} & \textbf{Log-Linear MAE} & \textbf{Scaling Laws MAE} & \textbf{All Features MAE} & \textbf{p-val (corrected)} \\
\midrule
\multicolumn{7}{c}{\textbf{Accuracy}} \\
\midrule
Arc Challenge  & 25-shot & 13.23\% & 4.91\% & \pmae{4.36}{0.12}  & \pmaeBS{3.67}{0.09}  & $3.67\times10^{-19}$ \\
GSM8k          & 5-shot  & 15.65\% & 11.03\% & \pmae{6.04}{0.21}  & \pmaeBS{5.10}{0.23}  & $5.41\times10^{-14}$ \\
Hellaswag      & 10-shot & 12.26\% & 4.29\% & \pmae{3.93}{0.13}  & \pmaeBS{3.18}{0.09}  & $4.44\times10^{-20}$ \\
Humaneval      & 0-shot  & 11.79\% & 8.61\% & \pmae{8.08}{0.22}  & \pmaeBS{6.93}{0.22}  & $4.44\times10^{-20}$ \\
Lambada        & 0-shot  & 16.89\% & 9.60\% & \pmae{9.51}{0.33}  & \pmaeBS{6.85}{0.25}  & $2.87\times10^{-22}$ \\
MMLU (0-shot)  & 0-shot  & 11.98\% & 9.12\% & \pmae{4.76}{0.20}  & \pmaeBS{4.10}{0.17}  & $5.26\times10^{-13}$ \\
MMLU (5-shot)  & 5-shot  & 12.25\% & 8.39\% & \pmae{3.97}{0.18}  & \pmaeBS{3.54}{0.14}  & $2.09\times10^{-10}$ \\
TruthfulQA     & 0-shot  &  3.72\% & 3.40\% & \pmae{2.75}{0.08}  & \pmaeBS{2.29}{0.06}  & $1.02\times10^{-17}$ \\
Winogrande     & 5-shot  & 10.14\% & 3.99\% & \pmae{3.39}{0.08}  & \pmaeBS{3.09}{0.07}  & $5.26\times10^{-13}$ \\
\midrule
\multicolumn{7}{c}{\textbf{Brier score}} \\
\midrule
XNLI           & 0-shot  & 7.22    & 5.45 & \mae{5.11}{0.11}  & \maeBS{4.30}{0.11}  & $1.37\times10^{-2}$ \\
ANLI           & 0-shot  & 9.48    & 5.95 & \mae{6.18}{0.19}  & \maeBS{5.86}{0.21}  & $3.16\times10^{-9}$ \\
MathQA         & 0-shot  & 7.57    & 3.89 & \mae{2.83}{0.06}  & \maeBS{2.75}{0.07}  & $1.63\times10^{-4}$ \\
LogiQA2        & 0-shot  & 12.62   & 8.77 & \mae{4.74}{0.12}  & \maeBS{4.60}{0.15}  & $3.84\times10^{-4}$ \\
\bottomrule
\end{tabular}
}
\caption{Comparison of MAE values (mean $\pm$ 95\% CI) for Scaling Laws and All Features predictors alongside Baseline MAE and Log-Linear MAE. Lower MAE is bolded; * indicates significance ($p<0.05$). Brier score values are multiplied by 100 to be on a similar scale to accuracy.}
\vspace{-1em}
\label{tab:comparison_final}
\end{table*}

\section{Results}
\subsection{Predictor Performance}
\label{sec:predictor_performance}

% \begin{table*}[!ht]
% \centering
% \small
% \begin{tabular}{ll c c c}
% \toprule
% \textbf{Benchmark} & \textbf{Setting} & \textbf{Baseline MAE} & \textbf{Scaling Laws MAE} & \textbf{All Features MAE} \\
% \midrule
% Arc challenge & 25 shot & 12.370\% & 5.400±2.030 & 3.690±1.380 \\
% Gsm8K & 5 shot & 20.300\% & 12.610±7.270 & 7.730±4.230 \\
% Hellaswag & 10 shot & 10.830\% & 4.490±0.730 & 3.040±0.370 \\
% Humaneval & 0 shot & 12.700\% & 10.220±6.330 & 7.730±4.960 \\
% Lambada & 0 shot & 16.510\% & 9.900±2.380 & 6.890±1.870 \\
% Mmlu & 0 shot_0 shot & 14.540\% & 10.880±8.040 & 8.970±7.440 \\
% Mmlu & 5 shot_5 shot & 15.780\% & 7.620±4.340 & 6.420±3.840 \\
% Truthfulqa & 0 shot & 3.690\% & - & - \\
% Winogrande & 5 shot & 9.280\% & 3.680±0.800 & 3.140±0.590 \\
% \bottomrule
% \end{tabular}
% \caption{Comparison of MAE values with linear predictor}
% \label{tab:comparison_final}
% \end{table*}

\textbf{Incorporating scale-independent features consistently improves benchmark performance.} We find that incorporating extra features alongside traditional scaling laws features leads to substantial improvements in prediction accuracy across multiple benchmarks, as seen in \autoref{tab:comparison_final}. The all-features predictor outperforms the scaling-laws-only predictor in all evaluated cases, with improvements ranging from approximately 3\% (MathQA) to about 28\% (Lambada) relative error reduction. Notably, the strongest improvements were observed in language modeling and common-sense reasoning tasks.

%This is convincing empirical evidence that across the 94 various models that we examine, it is \emph{not just scaling} that determines their downstream performance. 

\textbf{Certain tasks are more strongly dependent on non-scale features.} This pattern of improvements suggests that architectural and training data features may be more informative for predicting performance on certain types of tasks more strongly linked to a particular ``genre'' of data. Large improvements were observed for both code generation (HumanEval, $15\%$ improvement) as well as natural-language based reasoning tasks (e.g. Lambada, $28\%$ improvement). Even tasks with narrower domains, such as mathematical reasoning (GSM8k, +16\%) or knowledge-intensive evaluations (MMLU, +11–14\%), see consistent, if more moderate, enhancements. The Brier score benchmarks, however, show smaller improvements (around 3–6\%). This may be because the Brier score is inherently less sensitive to emergent effects in model performance, the specific choice of tasks limits the room for improvement, or a combination of both factors.%We discuss possible reasons for these differences in ($\S\ref{sec:what_features}$).

%\textbf{N-shot performance may be less strictly governed by scaling laws.} Our analysis of MMLU shows that the relative improvement is larger when few-shot examples are provided. For instance, in the 0-shot setting, the scaling-laws predictor yields a relative improvement of about 4.39\%. By contrast, in the 5-shot setting the relative improvement of approximately 6.09\%. A greater number of features are also selected as useful in the 5-shot case. However, since we have not systematically varied the number of few-shot examples across all benchmarks, this pattern is observational and may not hold universally.
%\gncomment{This paragraph is a candidate for removal if we need to save space.}

\begin{figure*}[!t]
    \centering
    % The image is displayed at full width:
    \includegraphics[width=0.9\linewidth]{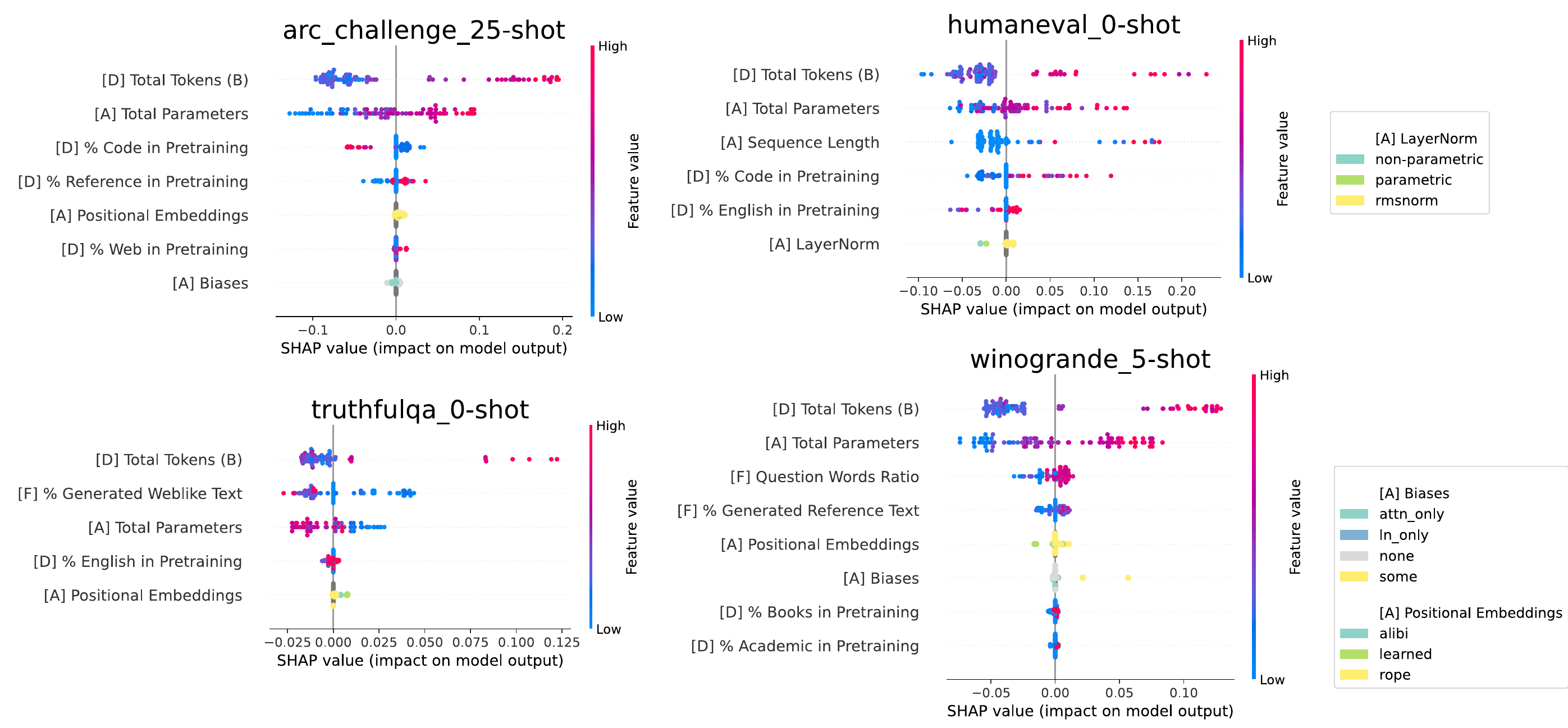}
    
    % Now, for the table, use adjustbox to ensure it doesn't exceed \textwidth,
    % and set its font size to \footnotesize:
    \begin{adjustbox}{max width=0.6\textwidth}
      {\footnotesize
      \begin{tabular}{lp{9cm}}
        \toprule
        \textbf{Feature Name} & \textbf{Description} \\
        \midrule
        {[D] Total Tokens (B)} & Total number of tokens used during pretraining, measured in billions (log scale). \\
        {[A] Total Parameters} & Total number of parameters in the model (log scale). \\
        {[D] \% Code in Pretraining} & Percentage of pretraining data that consists of code. \\
        {[F] Question Words Ratio} & Ratio of question-related words generated by the model. \\
        {[A] Dimension} & Embedding dimension. \\
        {[A] Sequence length} & Sequence length. \\
        {[A] LayerNorm} & Type of layer normalization used (non-parametric, parametric, rmsnorm). \\
        {[A] Biases} & Presence of bias parameters in the model. \\
        {[A] Positional Embeddings} & Type of positional embedding used (alibi, learned, rope). \\
        {[F] \% English Generated} & Percentage of English text generated by the model. \\
        {[D] \% Academic in Pretraining} & Percentage of pretraining data from academic sources. \\
        {[D] \% Reference in Pretraining} & Percentage of pretraining data from reference sources. \\
        {[F] \% Generated Weblike Text} & Percentage of web-like text generated by the model. \\
        {[F] \% Generated Reference Text} & Percentage of reference-like text generated by the model. \\
        {[D] \% Books in Pretraining} & Percentage of pretraining data from books. \\
        {[D] \% English in Pretraining} & Percentage of English text in the pretraining data.  \\
        \bottomrule
      \end{tabular}
      }
    \end{adjustbox}
    
    \caption{In all tasks, the number of parameters and pretraining tokens heavily influences the predictions made by the regressor. The percentage of code in pretraining often influences predictions negatively for NLI tasks but positively for Humaneval. [D], [A] and [F] denote features derived from data, architecture, or free-generations of a model respectively. \vspace{-1em}}
    \label{fig:shap_summary}
\end{figure*}

\begin{table*}[h]
    \centering

    %\caption{Description of extracted features used in the SHAP analysis.}
    \label{tab:feature_descriptions}
\end{table*}

\subsection{What Features Does Task Performance Depend On?}
\label{sec:what_features}

%To better understand the factors that influence task performance, we examine feature importances from XGBoost \gncomment{what do these represent? I think it's the number of trees in which the features are used, but I'm not absolutely certain}, 

To understand factors influencing task performance, we examine \citet{shap} (SHAP) values, which show how feature values affect predictions. Results for Arc Challenge, HumanEval, Winogrande, and TruthfulQA appear in \autoref{fig:shap_summary}, with remaining benchmarks in \autoref{appendix:all_shap_plots}.

\begin{figure}[ht]
    \centering
    \includegraphics[width=0.7\linewidth]{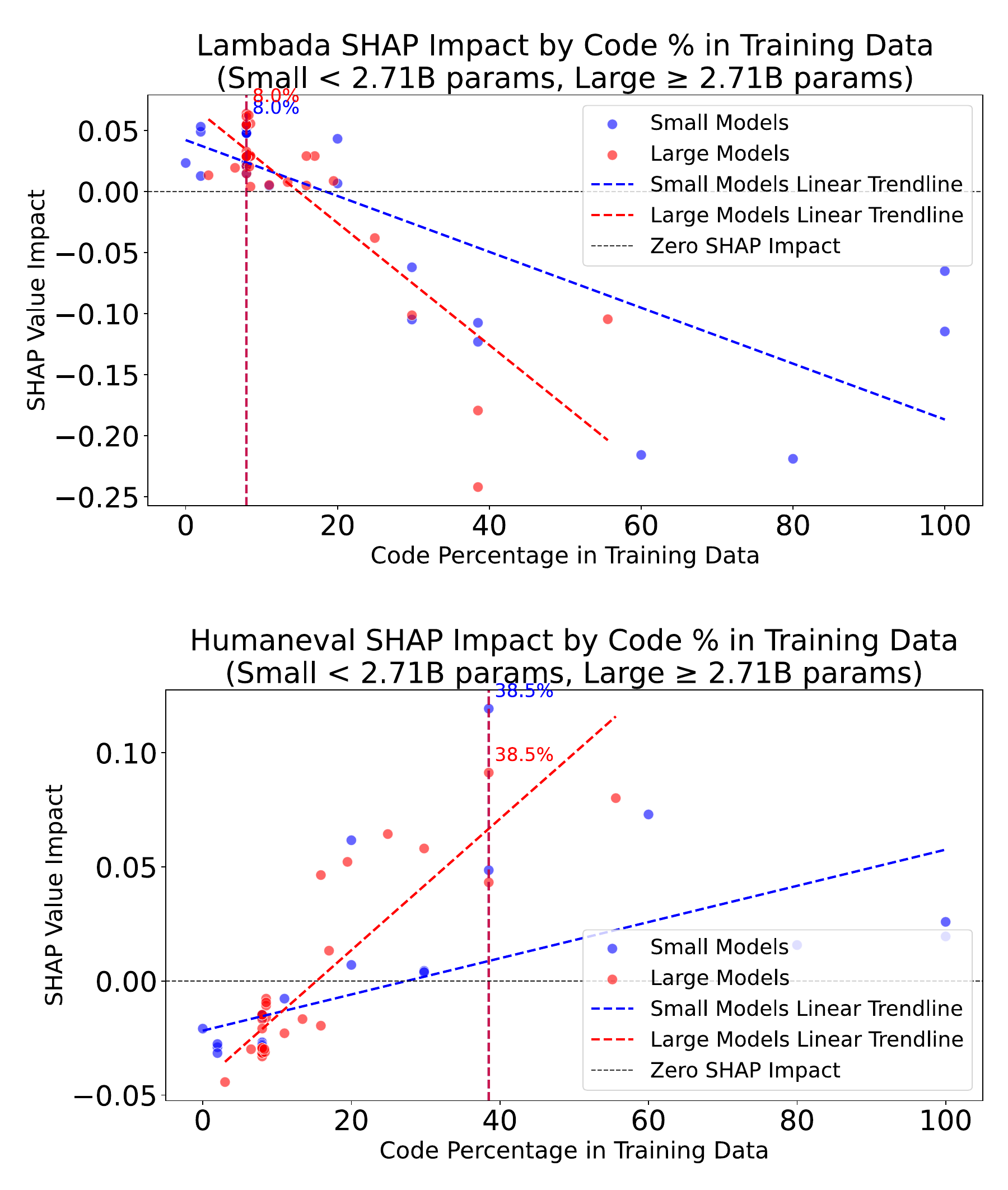}
    \caption{SHAP impact of code percentage on Lambada (reprentative NL task) and Humaneval on our regressors. \vspace{-1em}}
    \label{fig:code-nl-tradeoff}
\end{figure}

\textbf{A little code goes a long way, but too much is harmful to NLI.} The percentage of code in pretraining is a critical non-scaling feature. Higher code composition benefits Humaneval performance but is associated with lower scores on natural language reasoning tasks including Arc Challenge, Hellaswag, Winogrande, and Lambada. As shown in \autoref{fig:code-nl-tradeoff}, models with over 20-25\% code show gains on Humaneval but penalties on language benchmarks. A moderate 15-25\% code proportion appears to balance these competing demands.

% We emphasize that these conclusions stem from a relatively small sample of models, so more controlled experiments would be needed to confirm the exact “sweet spot.” Additionally, it is possible that large and small models may have slightly different scaling trends, though the lack of $>$3B models trained on very high percentages of code makes this difficult to verify.

\textbf{Other data domains show task-specific effects.} From free-generation features, we observed recent models trained on synthetic data (Phi \citep{textbooks_are_all_you_need}, SmolLM \citep{allal2024smollm}) generate more question words, suggesting training on question-answering content. Reference-like or question-loaded generations correlate with better performance on Arc Challenge and Winogrande, while web-like generations correlate with worse TruthfulQA performance (\autoref{fig:shap_summary}).

% However, empirical interventions may be needed to confirm whether selecting more question-oriented data or web data has the predicted effect.

\textbf{Non-scale architectural decisions have minor effects.}
Most highly influential features were data-related or architectural features related to scale (e.g., dimension).
However, both the type of layer norm and the positional embedding were deemed to have a significant effect in some cases.

\section{Validating Performance Predictions with Confirmatory Experiments}

\begin{figure}[!ht]
    \centering
    \includegraphics[width=0.8\linewidth]{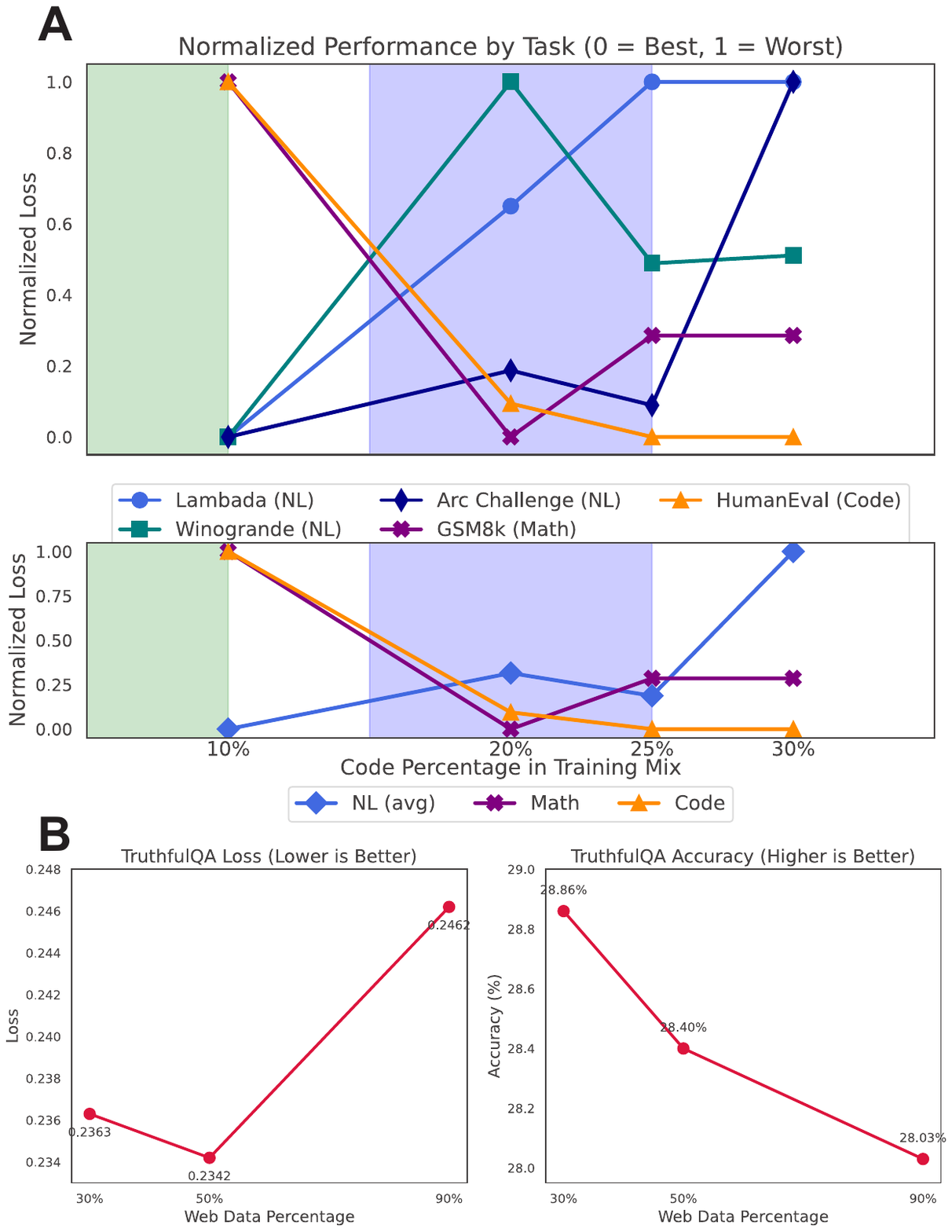}
    \caption{Loss on 460M parameter models, early checkpoints taken at 10\% of final training length. Panel A depicts the effects of code vs. natural language data on natural language datasets and coding, while Panel B depicts the effects of web vs. non-web based data on truthfulQA. In Panel A, the optimal ranges for natural language tasks only (green) and for balancing natural language and code (blue) are highlighted.}
    \label{figure:confirmatory_exps_figure}
\end{figure}

To validate findings from the meta-analysis, we also ran confirmatory pretraining runs with 460M-parameter models on the Dolma dataset. We aimed to validate two data distributional findings: (1) Around 8\% code is optimal when only considering natural language inference, but 15-25\% may be best when balancing code and natural language, and (2) TruthfulQA performance decreases with an increasing proportion of web data. As this is a small scale model and accuracy differences may not be significant, we convert the relevant datasets to use loss-based evaluations. Due to computational constraints, we train each checkpoint for 10B tokens, but use a cosine learning rate schedule scaled to a 100B token run. See \autoref{appendix:confirmatory_exps} for details and exact loss figures. Overall, in \autoref{figure:confirmatory_exps_figure} find that the confirmatory runs largely validated our meta-analysis predictions, with the exception that our margin-based loss for truthfulQA was slightly lower for the 50\% web data checkpoint compared to the 30\% checkpoint, though the trend for accuracy is as expected.
This provides preliminary evidence that our analysis method could be used to intelligently predict LM training design decisions a-priori.

% \begin{itemize}
%     \item{SHAP analysis description}
%     \item{Which features (besides scale) matter for each task: \% code for commonsense tasks and truthfulQA, \% reference for MMLU, architectural features}
%     \item{Caveat: architectural features may just be because newer models tend to use things like GQA which weren't a thing earlier}
%     \item{Ablating all scale related features: we still do better than median baselines, but see influence of non-scale related features more closely}
% \end{itemize}

% \section{Validation}
% \el{TODO: hopefully I can at least do a small experiment for this...}

% \begin{itemize}
%     \item{Hold everything fixed between 2 models, use median value for data distributional features}
%     \item{Plot predicted performance when varying one feature (the important ones from results sec)}
%     \item{See if trading off web vs code/reference has expected results}
%     \item{\el{Maybe use 1B models from regmix if really not enough time?}}
% \end{itemize}

% \begin{figure}[!htbp]
%   \centering
%   \fbox{%
%     \begin{minipage}[c][2in][c]{0.9\linewidth} % 2 inches tall, 90% of the line width
%       \centering
%       Placeholder for predicted vs. actual performance for a 220M/1B model on benchmarks
%     \end{minipage}%
%   }
%   \caption{Placeholder for predicted vs. actual performance for a 220M/1B model on benchmarks}
%   \label{fig:placeholder}
% \end{figure}

%\subsection{Model Training}

\section{Publication Bias Sensitivity Check}

One other potential caveat to observational studies is that positive results are more likely to be reported: especially in the case of architecture, people are likely to release a new model when the model does well. There are two cases of publication bias, using architecture as an example. There are two sources of publication bias, first in design space: some architectural variants that perform particularly poorly or well may never be trained or reported; these are inherently unobservable and we cannot comment on them. We focus on the second source of bias (small-scale bias), which is that among the results that we \textit{do} observe, noisier task-level estimates are more likely to look large, and thus be emphasized. We focus on this type of bias correction, which is often examined in meta-analyses.

For each default benchmark/setting in our main suite, we computed a task-level contrast (one-vs-rest for a given architectural level, e.g. RoPE vs non-RoPE positional embeddings) via precision-weighted WLS with controls for log parameters and log pretraining tokens. We then meta-regressed these task-level effects $y_i$ on their standard errors $SE_i$ via a standard method known as PET-PEESE (\citealt{StanleyDoucouliagos2014PETPEESE}). Across the default \emph{accuracy} tasks ($k\approx6$ per feature), the pooled, bias-corrected effects are small: RoPE $\approx\!+0.015$, RMSNorm $\approx\!+0.023$, GQA $\approx\!+0.034$ (with wider uncertainty), while other levels are near zero. Results can be found in \autoref{appendix:publication_bias}. These adjustments leave our qualitative story unchanged: non-scale architectural choices have, on average, modest benchmark-robust effects compared to data composition and scale.

\section{Related Work}

%\gncomment{I haven't read this carefully, but it can probably be compressed a little.}

\subsection{Empirical Data Composition Results}
Prior work has examined code in pretraining \cite{ma2023trainingstagedoescode, aryabumi2024codecodeexploringimpact} and domain ablations \cite{longpre2023pretrainer}. Data filtering improves performance beyond scaling alone \cite{sorscher2023neuralscalinglawsbeating, goyal2024scalinglawsdatafiltering}. Our results show code enhances natural language reasoning at moderate proportions (optimal ratio 15-25\%), refining previous estimates of 25\% \cite{aryabumi2024codecodeexploringimpact}. Our approach of pooling insights from existing models complements empirical ablations by identifying promising axes for further testing.

\subsection{Observational and Task-Specific Scaling Law Fitting}
\label{sec:rw-other-scaling-laws}
Task-specific scaling laws research shows parameter allocation affects machine translation outcomes \cite{ghorbani2021scalinglawsneuralmachine}, and multitasking benefits English-target languages \cite{fernandes23a}. Work on downstream tasks emphasizes alignment between pretraining and downstream data \cite{hernandez2021scaling, isik2024scaling}. Various studies address data repetition \cite{muennighoff2024scaling}, multiple domains \cite{goyal2024scalinglawsdatafiltering}, and factors like sparsity \cite{frantar2023scaling}, precision \cite{kumar2024scalinglawsprecision}, and inference costs \cite{hoffmann2022training}, while some find stability across training hyperparameters \cite{deepseekai2024deepseekllmscalingopensource}.

\citet{ruan2024observationalscalinglawspredictability} also use observations from open-source models to predict task performance, but derive their predictions of one task's performance from performance on other tasks. We find a similar result in identifying two axes of performance-- general natural language ability and coding ability but are motivated instead by tracing these capabilities back to pretraining decisions.

\subsection{Pretraining Data Selection}

Domain mixing has been studied in pretraining, and other works have formulated this as a regression problem \cite{ ye2024datamixinglawsoptimizing, liu2025regmixdatamixtureregression} or used proxy models to select domain weights in the course of training \cite{xie2023doremioptimizingdatamixtures, albalak2023efficientonlinedatamixing, jiang2024adaptivedataoptimizationdynamic, yu2025dataefficientpretraininggroupleveldata}. In contrast, we retrospectively analyze how domain composition and training decisions influence performance across tasks, which is a complementary perspective to optimizing data weights for a single model during training.

\subsection{Tracing Capabilities to Data}

Specific language model capabilities have been linked to patterns in pretraining data. Performance on numerical reasoning and syntactic rule learning depends on frequency of numerical terms in the training data \citep{kassner2020pretrained, wei2021frequency}. \citet{ruis2024proceduralknowledgepretrainingdrives} found that influential data for reasoning is dispersed across numerous documents and is associated with procedural content. Similarly, \citet{chen2024parallelstructurespretrainingdata} observed that "parallel structures" are closely tied to in-context learning abilities. We currently focus on broader data domains, but our framework can be extended with more granular tasks or refined data features.

\section{Conclusion and Future Work}
\label{sec:conclusion}

We perform the first systematic analysis of the performance of open language models across diverse tasks and tie their performance to architectural and data-compositional design decisions.
Looking into the future, there are a number of clear directions.
First, our database ($\S\ref{sec:database}$) can be further expanded as new models and benchmarks are released, and we will release the code and data to help spur community efforts for more systematic data documentation.
Second, we hope our work will help discover hypotheses to be tested in more controlled settings -- existing models intertwine a number of design decisions, and further controlled pre-training experiments that only involve one axis of variation could further clarify the effect of each feature.
Finally, within our study, the great majority of pre-trained models focused on dense transformer architectures, while alternative architectures such as mixture-of-experts \cite{jiang2024mixtralexperts,deepseekai2024deepseekv3technicalreport} and state-space models \cite{mamba} have also seen significant research interest.
How to appropriately featurize these more various model architectures and use the information in performance prediction is an interesting challenge that may uncover further insights. Lastly, although pretraining data analysis and selection has mainly been focused on empirical findings so far, building a better understanding of how training affects model capabilities through large-scale empirical studies could also facilitate interpretability experiments and possible interventions on learned representations, with controlled axes of variation providing case studies.

\newpage
\section*{Limitations}
Our current work has several limitations that can be improved in future work. First, although we document many open models, our sample size remains limited, particularly for larger ($>$50B) parameter models. This limits our ability to draw robust conclusions about scaling behaviour in large models. Additionally, the models that we have are not evenly distributed across number of parameters, data size, and data distributions, with certain size ranges and data distributions being overrepresented. There are also likely selection effects in which models are made open-weights, as well as likely time effects in popular architectural decisions or data compositions in different time periods.

Second, our methodology also imposes some limitations. Because we do not systematically train all our own models (though we have a few of our own in \autoref{appendix:all_models}), our analyses are observational in nature. While we can observe interesting relationships between design choices and performance, making causal claims requires experimental validation. Additionally, while tree-based regressors are effective for capturing complex feature interactions, they limit our ability to extrapolate beyond the range of model sizes (in parameters and tokens) seen in our dataset.

Last, we note that the scope of our work also has limitations. Namely, we focus on base pretrained decoder-only dense transformer models, which excludes significant architectural variants such as mixture-of-experts models, non-transformer based architectures, as well as post-trained models. Additionally, we examine mostly English-language models as we do not focus on multilinguality in this work. Our feature set, while extensive, may also not capture all relevant details of model design and training, particularly optimization details as of now.

These limitations suggest directions for future work: expanding the database to include more diverse model types and language coverage, developing more targeted functional forms that allow better extrapolation while also taking as input a heterogeneous feature set, as well as conducting targeted experiments with new pretrained models to validate the impact of specific design choices.
% We find that augmenting traditional scaling features with scale-independent features, especially data composition related features, consistently improves prediction of LM performance. Our findings offer several concrete takeaways for model design. First, while code data boosts performance on tasks like code generation, too high a percentage harms natural language understanding. A balanced approach—keeping code content in the 10–20\% range—appears optimal. Second, including a modest amount of multilingual data can enhance performance on certain English tasks, suggesting that a diversified pretraining mix is beneficial. Finally, the impact of these scale-independent features may be more pronounced in few-shot settings, suggesting that evaluating models with more few-shot examples may better distinguish the effects of data distributions. We hope that by systematically documenting training decisions and their impact, this resource can serve to guide refinements in language model training.

\section*{Ethical Considerations}

In this work, we focus on understanding why models may perform well on standard benchmarks, but do not focus on other important considerations such as safety or societal bias.

Furthermore, our analysis focuses on English-language models and benchmarks. This limitation reflects but may also reinforce the field's existing bias toward English, potentially contributing to underinvestment in developing effective architectures for other languages.

\section*{Author Contribution Statement}

We follow a slightly modified version of the CRediT author statement. 
\begin{itemize}
    \item \textbf{EL}: conceptualization (lead), data curation (equal), methodology/software (lead), writing -- original draft (lead), writing -- review and editing (lead), visualization (lead), funding acquisition (supporting)
    \item \textbf{AB}: conceptualization (supporting), methodology/software (supporting), data curation (equal), writing -- review and editing (supporting)
    \item \textbf{LS}: conceptualization (supporting), methodology/software (supporting), data curation (equal)
    \item \textbf{LT}: conceptualization (supporting), methodology/software (supporting), data curation (equal)
    \item \textbf{PF}: conceptualization (supporting), methodology/software (supporting), data curation (equal), writing -- review and editing (supporting)
    \item \textbf{LM}: methodology/software (supporting), data curation (equal)
    \item \textbf{MC}: methodology/software (supporting), data curation (equal)
    \item \textbf{SS}: methodology/software (supporting), data curation (equal)
    \item \textbf{CL}: writing -- review and editing (supporting), project administration (supporting)
    \item \textbf{AR}: conceptualization (supporting), writing -- review and editing (supporting), supervision (supporting)
    \item \textbf{KG}: conceptualization (supporting), writing -- review and editing (supporting), supervision (supporting), project administration (supporting)
    \item \textbf{GN}: conceptualization (lead), methodology/software (supporting), writing -- review and editing (supporting), supervision (lead), project administration (lead), funding acquisition (lead)
\end{itemize}

\section*{Acknowledgments}

This work was supported by a fellowship from NEC Laboratories Europe and the BRIDGE CMU-AIST research project. EL was supported by the Natural Science and Engineering Research Council of Canada (NSERC), [funding reference number 578085]. AB was supported by a grant from the National Science Foundation Graduate Research Fellowship Program under Grant No. DGE2140739.
Any opinions, findings, and conclusions or recommendations expressed in this material are those of the author(s) and do not necessarily reflect the views of the sponsors.

\section*{Use of AI Assistants}

Claude 3.5 Sonnet and GPT-o3-mini-high were used to help revise and shorten several parts of this submission as well as to edit for clarity. The first draft was entirely human-written. 

% help -- tbh I forgot who I discussed this with but adding people who edited
%We thank Xiang Yue for providing useful feedback and advice on this work, as well as Simran Khanuja for providing helpful edits. We also thank Brendon Boldt and David Mortensen for help with \LaTeX \hspace{0.2mm} formatting.

\bibliography{acl_latex}

\appendix

\section{List of all models}
\label{appendix:all_models}

All models are listed in \autoref{tab:all_models}.

\onecolumn
\begin{longtable}{p{0.35\textwidth}p{0.45\textwidth}r}
\caption{Model Parameter Counts by Organization (sorted by size)} \\
\label{tab:all_models} \\
\toprule
Organization & Model Name & Parameters \\
\midrule
\endfirsthead

\multicolumn{3}{c}{\tablename\ \thetable\ -- Continued from previous page} \\
\toprule
Organization & Model Name & Parameters \\
\midrule
\endhead

\midrule
\multicolumn{3}{r}{Continued on next page} \\
\endfoot

\bottomrule
\endlastfoot
EleutherAI \cite{pythia} & pythia-14m & 14M \\
EleutherAI & pythia-70m-deduped & 70M \\
EleutherAI & pythia-70m & 70M \\
facebook \cite{opt} & opt-125m & 125M \\
EleutherAI \cite{gptneo} & gpt-neo-125m & 125M \\
HuggingFaceTB \cite{allal2024smollm} & SmolLM-135M & 135M \\
EleutherAI & pythia-160m & 160M \\
EleutherAI & pythia-160m-deduped & 160M \\
None (this paper) & llama2\_220M\_nl\_100\_code\_0 & 220M \\
None (this paper) & llama\_220M\_nl\_80\_code\_20 & 220M \\
None (this paper) & llama2\_220M\_nl\_40\_code\_60 & 220M \\
None (this paper) & llama2\_220M\_nl\_20\_code\_80 & 220M \\
None (this paper) & llama2\_220M\_nl\_0\_code\_100 & 220M \\
Salesforce \cite{codegen} & codegen-350M-mono & 350M \\
Salesforce & codegen-350M-multi & 350M \\
Salesforce & codegen-350M-nl & 350M \\
facebook & opt-350m & 350M \\
HuggingFaceTB & SmolLM-360M & 360M \\
EleutherAI & pythia-410m-deduped & 410M \\
EleutherAI & pythia-410m & 410M \\
facebook \cite{xglm} & xglm-564M & 564M \\
EleutherAI & pythia-1b-deduped & 1B \\
bigscience \cite{bloom} & bloom-1b7 & 1B \\
EleutherAI & pythia-1b & 1B \\
cerebras \cite{cerebras} & Cerebras-GPT-1.3B & 1.3B \\
microsoft \cite{phi1_5} & phi 1.5 & 1.3B \\
EleutherAI & gpt-neo-1.3B & 1.3B \\
EleutherAI & pythia-1.4b & 1.4B \\
EleutherAI & pythia-1.4b-deduped & 1.4B \\
HuggingFaceTB & SmolLM-1.7B & 1.7B \\
Salesforce & codegen-2B-mono & 2B \\
Salesforce & codegen-2B-nl & 2B \\
Salesforce & codegen-2B-multi & 2B \\
google \cite{gemma2} & gemma-2-2b & 2B \\
cerebras & Cerebras-GPT-2.7B & 2.7B \\
EleutherAI & gpt-neo-2.7B & 2.7B \\
NinedayWang \cite{polycoder} & PolyCoder-2.7B & 2.7B \\
facebook & opt-2.7b & 2.7B \\
microsoft \cite{phi2} & phi 2 & 2.7B \\
EleutherAI & pythia-2.8b & 2.8B \\
EleutherAI & pythia-2.8b-deduped & 2.8B \\
facebook & xglm-2.9B & 2.9B \\
Qwen \cite{qwen_2-5} & Qwen2.5-3B & 3B \\
cerebras \cite{btlm} & btlm-3b-8k-base & 3B \\
openlm-research \cite{openllama} & open\_llama\_3b\_v2 & 3B \\
rinna \cite{sawada2024release} & bilingual-gpt-neox-4b & 4B \\
Dampish & StellarX-4B-V0 & 4B \\
facebook & xglm-4.5B & 4.5B \\
Salesforce & codegen-6B-multi & 6B \\
EleutherAI \cite{gptj} & gpt-j-6b & 6B \\
Salesforce & codegen-6B-nl & 6B \\
Salesforce & codegen-6B-mono & 6B \\
cerebras & Cerebras-GPT-6.7B & 6.7B \\
facebook & opt-6.7b & 6.7B \\
EleutherAI & pythia-6.9b-deduped & 6.9B \\
EleutherAI & pythia-6.9b & 6.9B \\
Qwen \cite{qwen1} & Qwen-7B & 7B \\
aisingapore \cite{lowphansirikul2021wangchanberta} & sea-lion-7b & 7B \\
bigscience & bloom-7b1 & 7B \\
google \cite{gemma1} & gemma-7b & 7B \\
mosaicml \cite{mpt} & mpt-7b & 7B \\
openlm-research & open\_llama\_7b & 7B \\
tiiuae \cite{falcon} & falcon-7b & 7B \\
allenai \cite{olmo} & OLMo-7B-hf & 7B \\
huggyllama \cite{touvron2023llama} & llama-7b & 7B \\
LLM360 \cite{liu2023llm360fullytransparentopensource} & Amber & 7B \\
LLM360 & CrystalCoder & 7B \\
facebook & xglm-7.5B & 7.5B \\
meta-llama \cite{llama3} & Meta-Llama-3-8B & 8B \\
google & gemma-2-9b & 9B \\
01-ai \cite{yi} & Yi-9B & 9B \\
EleutherAI & pythia-12b & 12B \\
EleutherAI & pythia-12b-deduped & 12B \\
cerebras & Cerebras-GPT-13B & 13B \\
meta-llama \cite{llama2} & Llama-2-13b-hf & 13B \\
Qwen & Qwen1.5-14B & 14B \\
Qwen & Qwen2.5-14B & 14B \\
Salesforce & codegen-16B-nl & 16B \\
Salesforce & codegen-16B-mono & 16B \\
EleutherAI & gpt-neox-20b & 20B \\
mosaicml & mpt-30b & 30B \\
Qwen & Qwen2.5-32B & 32B \\
Qwen & Qwen1.5-32B & 32B \\
AbacusResearch & Jallabi-34B & 34B \\
01-ai & Yi-34B & 34B \\
01-ai & Yi-34B-200K & 34B \\
meta-llama & Llama-2-70b-hf & 70B \\
meta-llama \cite{llama3} & Meta-Llama-3.1-70B & 70B \\
meta-llama & Meta-Llama-3-70B & 70B \\
Qwen \cite{qwen_2-5} & Qwen2-72B & 72B \\
Qwen & Qwen2.5-72B & 72B \\
Qwen & Qwen1.5-110B & 110B \\
\end{longtable}

%Model citations: \cite{touvron2023llama, pythia, codegen, cerebras, mosaicmlmpt, falcon, olmo, opt, xglm, llama3}

\twocolumn
\section{List of all architectural and data features}
\label{appendix:all_features}

\subsection{Architectural Features}

Note that features in this section are collected from official documentation (e.g. huggingface model/data cards or original papers).

\begin{itemize}
    \item \textbf{Total parameters} - the total number of parameters (embedding included) in the model. Note that we only include decoder-only dense models.
    \item \textbf{Dimension} - the embedding dimension.
    \item \textbf{Num heads} - the number of attention heads.
    \item \textbf{MLP ratio} - the ratio of $\frac{\text{FFN dimension}}{\text{embedding dimension}}$.
    \item \textbf{Positional Embeddings} - the type of positional embedding. This is either non-parametric (sinusoidal or fixed embeddings), learned (just learned as a vector per position), rope (rope embeddings), or alibi (technically not an embedding, but included here due to its functional purpose)
    \item \textbf{LayerNorm} - the type of layernorm applied. This is either non-parametric (just an arithmetic based normalization), parametric (similar, but with some learnable parameters such as scaling/biases), and RMSNorm (a simplified version of parametric)
    \item \textbf{Attention variant} - The broad type of attention used. This is either full (vanilla attention), local (each token position only attends to positions around it), mqa (multi-query attention), or gqa (grouped-query attention)
    \item \textbf{Biases} - whether or not bias terms are present in parts of the model. Either none (no biases), attn only (only in attention layers), ln only (only in layer norm)
    \item \textbf{Block type} - whether or not the transformer blocks are computed in parallel at all. Sequential indicates not, while parallel indicates some parallelism in attention or FFN layers. 
    \item \textbf{Activation} - the activation function used. Either relu, gelu/gelu variations, silu, or swiglu. 
    \item \textbf{Sequence length} - the sequence length.
    \item \textbf{Batch instances} - the batch size used during pretraining.
\end{itemize}

\subsection{Data Features}

Note that features in this section are collected from official documentation (e.g. huggingface model/data cards or original papers).

\begin{itemize}
    \item \textbf{Total tokens (B)} - total number of tokens used during pretraining, measured in billions (converted to log scale)
    \item \textbf{\% Web in Pretraining} - Percentage of pretraining data from general web sources.
    \item \textbf{\% Code in Pretraining} - Percentage of pretraining data that consists of code.
    \item \textbf{\% Books in Pretraining} - Percentage of pretraining data from books.
    \item \textbf{\% Reference in Pretraining} - Percentage of pretraining data from reference sources.
    \item \textbf{\% Academic in Pretraining} - Percentage of pretraining data from academic sources.
    \item \textbf{\% English in Pretraining} - Percentage of English text in the pretraining data.
\end{itemize}

\subsection{Freegen-derived Features}
These features are derived from model generations. For each model, 5--10k generations are extracted and the following metrics are aggregated (by mean and standard deviation). However, bigram entropy, the educational classifier score, and domain classifications are exceptions, as they are computed once across all generations.

We use Stanza \cite{stanza} to generate the parse-based features after classifying generations by language. We only include languages that are supported by stanza in the final set of generations that the parse features are based on.

\subsubsection{Generation Length \& Basic Statistics}
\begin{itemize}
    \item \textbf{Mean Character Length} -- Average number of characters per generation (capped at 2048).
    \item \textbf{Mean Tokens Generated} -- Average number of tokens per generation.
    \item \textbf{Mean Sentences} -- Average number of sentences per generation.
    \item \textbf{Mean Words} -- Average number of words per generation.
    \item \textbf{Mean Words per Sentence} -- Average number of words per sentence.
\end{itemize}

\subsubsection{Constituency Parse Features}
\begin{itemize}
    \item \textbf{Mean Depth of Deepest Parse Tree} -- Average maximum constituency tree depth per generation.
    \item \textbf{Mean Depth of Parse Trees} -- Average constituency tree depth across all sentences/phrases.
    \item \textbf{Mean Word Depth} -- Average depth of words within constituency trees.
    \item \textbf{Mean Word Depth Variation} -- Average standard deviation of word depths across sentences/phrases.
\end{itemize}

\subsubsection{Dependency Parse Features}
\begin{itemize}
    \item \textbf{Mean 90th-Percentile Dependency Head Distances} -- For each generation, compute the 90th-percentile of the linear distances between words and their dependency head, then average these values.
    \item \textbf{Mean Maximum Dependency Head Distances} -- Average maximum distance from any word to its dependency head per generation.
    \item \textbf{Mean Median Dependency Head Distances} -- Average median dependency-head distance per generation.
    \item \textbf{Mean Maximum Dependency Root Distances} -- Average maximum distance from any word to the sentence root per generation.
    \item \textbf{Mean Mean Dependency Root Distances} -- Average of the mean distances from words to the sentence root per generation.
    \item \textbf{Mean Median Dependency Root Distances} -- Average of the median distances from words to the sentence root per generation.
\end{itemize}

\subsubsection{Domain Classification Features}
\begin{itemize}
    \item \textbf{\% Generated Academic-like Text} -- Percentage of generations classified as academic-like.
    \item \textbf{\% Generated Books-like Text} -- Percentage of generations classified as books-like.
    \item \textbf{\% Generated Code-like Text} -- Percentage of generations classified as code-like.
    \item \textbf{\% Generated Reference-like Text} -- Percentage of generations classified as reference-like.
    \item \textbf{\% Generated Specialized Text} -- Percentage of generations classified as specialized (e.g., music scores, chess PGNs, biomedical data).
    \item \textbf{\% Generated Web-like Text} -- Percentage of generations classified as web-like.
\end{itemize}

\subsubsection{Classifier and Language Metrics}
\begin{itemize}
    \item \textbf{Mean Educational Classifier Score} -- Average score assigned by the educational classifier.
    \item \textbf{\% Generated English Text} -- Average percentage of text generated in English.
\end{itemize}

\subsubsection{Lexical Diversity and Entropy Metrics}
\begin{itemize}
    \item \textbf{Mean Bigram Entropy} -- Average entropy computed on bigrams across generations.
    \item \textbf{Type-Token Ratio} -- Average ratio of unique tokens to total tokens.
    \item \textbf{Unique Tokens} -- Average number of unique tokens per generation.
\end{itemize}

\subsubsection{Lexical and Stylistic Features}
\begin{itemize}
    \item \textbf{Content-Function Ratio} -- Ratio of content words (nouns, verbs, adjectives, adverbs) to function words.
    \item \textbf{Question Words Ratio} -- Ratio of question-related words (e.g. how, what, why, when, where, who, which, whose) per 100k words.
    \item \textbf{Imperative Words Ratio} -- Ratio of imperative words (e.g. do, make, consider, take, use, ensure, check, build, apply, run, create, find, go, try, turn, start, stop, put, keep, leave, get, move) per 100k words.
    \item \textbf{Conjunctions Ratio} -- Ratio of conjunction words (e.g. and, but, or, so, because, although, however, therefore, yet) per 100k words.
    \item \textbf{Instruction Words Ratio} -- Ratio of instruction-oriented phrases (e.g. “Question:”, “Answer:”, “Instruction:”, “User:”, “Assistant:”, “Q:”, “A:”) per 100k words.
    \item \textbf{Numbers Ratio} -- Ratio of numerical tokens in the generated text.
\end{itemize}

\section{List of all evaluations and settings}
\label{appendix:all_evals}

Although we ideally would evaluate the full cross-product of models and tasks, we found that due to some models being incompatible with LM Evaluation Harness and compute constraints we could not evaluate all \nummodels models on every dataset. We list in \autoref{tab:evals-progress} the number of evaluations we currently have for each benchmark and will continue to fill out evaluations in the database.

\begin{table}[t]
\centering
\renewcommand{\arraystretch}{0.9} % reduce row spacing
\resizebox{0.9\columnwidth}{!}{
\begin{tabular}{l c}
\toprule
\textbf{Task Name} & \textbf{\# Models Evaluated} \\
\midrule
\multicolumn{2}{c}{\textbf{Commonsense Reasoning / NLI}} \\
\midrule
ANLI & 82 \\
HellaSwag & 92 \\
Winogrande & 92 \\
XNLI & 82 \\
\midrule
\multicolumn{2}{c}{\textbf{Math / Logic}} \\
\midrule
GSM8K & 92 \\
LogiQA2 & 82 \\
MathQA & 82 \\
\midrule
\multicolumn{2}{c}{\textbf{General Knowledge}} \\
\midrule
ARC Challenge & 92 \\
Lambada & 92 \\
MMLU & 92 \\
\midrule
\multicolumn{2}{c}{\textbf{Other}} \\
\midrule
TruthfulQA & 92 \\
HumanEval & 91 \\
\bottomrule
\end{tabular}
}
\caption{Number of models evaluated for each benchmark task. Note that some models encountered technical errors when being loaded or in lm-eval-harness. The number of models will continue to be updated.}
\label{tab:evals-progress}
\end{table}

\section{Task Deviations from Kaplan-style Scaling Laws}
\label{appendix:task-deviation}

In \autoref{tab:pl_deviations}, we document the $R^2$ value for a fitted power law on the performance of each model. 

\begin{table}[ht]
\centering
\begin{tabular}{l c}
\toprule
\textbf{Benchmark} & \textbf{\(R^2\)} \\
\midrule
gsm8k & 0.85 \\
arc challenge & 0.82 \\
hellaswag & 0.80 \\
winogrande & 0.80 \\
mmlu 5-shot & 0.80\\
mmlu 0-shot & 0.74 \\
mathqa & 0.70 \\
ANLI & 0.61 \\
humaneval & 0.61 \\
lambada & 0.55 \\
LogiQA2 & 0.50 \\
XNLI & 0.41 \\
truthfulqa & 0.29 \\
\bottomrule
\end{tabular}
\caption{Overview of \(R^2\) values by benchmark. }
\label{tab:pl_deviations}
\end{table}

\section{Free-generation Domain Classification}
\label{appendix:freegen-prompt}

We classify model generations into top-level domains with GPT-4o-mini. We found that this multi-stage prompt \autoref{lst:prompt}, \autoref{lst:prompt_2} had reasonable precision on a sample of Dolma by domain \cite{dolma}, so use it to classify free-generations.

\begin{figure*}[t]
    \centering
    \begin{minipage}{\textwidth}
        \small % Adjust font size as needed
        \lstset{
            basicstyle=\ttfamily,
            breaklines=true,
            columns=fullflexible
        }
\begin{lstlisting}[caption={Multistage classification prompt.}, label={lst:prompt}, escapeinside={(*@}{@*)}]

<PROMPT 1>. [SYSTEM] You are a system tasked with classifying documents. First, determine if this document is relatively coherent. These documents are generated by language models, so they may not make sense. Classify a document as incoherent ONLY if it shows extreme repetition, code mixes in a way that does not make sense (such as different languages referencing entirely different subjects), or if it is mostly gibberish. Don't worry about logic errors or factual inconsistencies. If multiple documents are mixed into one, classify it as incoherent. Respond ONLY with "incoherent" if the document is incoherent, otherwise respond with "not_incoherent"
[USER] Please classify the document as incoherent or not_incoherent.\nDocument: {document}

If not incoherent...
<PROMPT 2>. [SYSTEM] Determine if this document contains programming code. Look for:
1. Programming language keywords (def, class, import, etc)
2. Code blocks (marked with backticks, indentation patterns)
3. Stack Overflow-style Q&A about programming
4. File extensions (.py, .js, etc)
5. Documentation about code/config files

Respond ONLY with:
- "code" if ANY of these are present
- "not_code" otherwise
[USER] Please classify the document as code or not_code.\nDocument: {document}

If not code...
<PROMPT 3> [SYSTEM] For documents WITHOUT programming code, determine if this is web content. Web content includes news articles, social media and online forums, blog posts, shopping websites, and other general websites. This includes a wide variety of content, and anything that looks like it may be a web article at all should be included. Look for:
1. URLs or hyperlinks
2. Social media formatting (@mentions, #hashtags)
3. "Click here" or UI elements
4. Comment threads or forum posts
5. Shopping/e-commerce language
6. Bylines or author names
7. Descriptions of products or product features

Respond ONLY with:
- "web" if ANY of these are present
- "not_web" otherwise

[USER] Please classify the document as web or not_web.\nDocument: {document}

If not web...
<PROMPT 4> [SYSTEM] For documents WITHOUT programming code, determine if this is academic or patent-related content. Academic content consists of research papers and snippets of research in both sciences and humanities, as well as patent applications. Student essays or assignments should also be included in this category. Look for:
1. Citations or references 
2. Latex formatting such as equations or tables
3. Formal academic language, not aimed at educating a general audience
4. Technical jargon or domain-specific terminology
5. Patent numbers or legal language (but not court documents, only patents)

Do NOT classify as academic if the document:
- Only uses occasional technical terms
- Is a popular science article or description of a scientific study, rather than the study itself
- Is educational but aimed at a general audience

Respond ONLY with:
- "academic" if ANY of these are present
- "not_academic" otherwise
[USER] Please classify the document as academic or not_academic.\nDocument: {document}
        \end{lstlisting}
    \end{minipage}
\end{figure*}

\begin{figure*}[t]
    \centering
    \begin{minipage}{\textwidth}
        \small % Adjust font size as needed
        \lstset{
            basicstyle=\ttfamily,
            breaklines=true,
            columns=fullflexible
        }
\begin{lstlisting}[caption={Multistage classification prompt (contd).}, label={lst:prompt_2}, escapeinside={(*@}{@*)}]
If not academic...
<PROMPT 5> [SYSTEM] For documents WITHOUT programming code, determine if this is a book, reference material (including media content), or a specific dataset. Books include literary works, fiction, and narrative nonfiction. Reference material includes wikipedia, dictionaries, textbooks and textbook like content, and encyclopedias. Please note that reference should also include instruction or human preference datasets for language model training. Media content includes podcasts, subtitles, and other media-related text. Specific datasets are unique and not covered by the other categories, such as biomedical datasets or molecules, chess PGNs or specific data formats not covered by any other category. Look for:
            
For the books category:
1. Chapter headings or book titles
2. Fictional character names or dialogue
3. For literary nonfiction, look for a more narrative and less didactic tone
4. Extended narrative prose or dialogue
Do NOT classify as books if the document:
- Only has a single dialogue snippet
- Could be a web article
- Is primarily informational or educational (use reference instead)

For the reference category:
1. Definitions or explanations of terms
2. Encyclopedic formatting
3. Textbook-like language
4. Explanations or examples meant to educate a reader
5. Chat formatting like 'User:/Assistant:' or similar tokens
6. Court documents or legal language (NOT patents)
7. Wikipedia headers such as 'references' or 'external links'

For the media category (should be classified as reference):
1. Audio or video timestamps
2. Subtitles or captions

For the specific datasets category:
1. Unique names or identifiers
2. Dataset-specific formatting
3. Data or metadata descriptions

If this seems to be a web document (social media, news, blogs, forums, shopping), you can also back off to the 'web' category.

Respond ONLY with:
- "books" if the document is a book
- "reference" if the document is reference material
- "specific_datasets" if the document is a specific dataset
- "web" if the document is web content
- "unknown" if none of these are present

[USER] Please classify the document as books, reference, media, specific_datasets, or unknown.\nDocument: {document}"
        \end{lstlisting}
    \end{minipage}
\end{figure*}

\section{Domain Classifier Validation}
\label{appendix:classifier_validation}

To validate the reliability of the 4o-mini based classifier, we had an author of this paper annotate 300 selected samples from three pretraining datasets (the Pile, the SmolLM corpus, and RefinedWeb) according to the same annotation standards used in \autoref{appendix:freegen-prompt}. Samples annotated as "unknown" or "incoherent" by either the model or the human annotator were excluded, as these samples are not included in computing the domain mix.

After filtering, we analyzed 258 text samples and found that the human annotator and the model had an 85.8\% absolute agreement, and a Cohen's $\kappa$ of 0.746, indicating high agreement between human classifications and the model's classifications.

\section{Free-generation Validation}
\label{appendix:freegen_validation}

To validate our free-generation approach as a proxy for pretraining data composition, we analyzed correlations between the free-generation features of models and their pretraining data. 

Namely, for models trained on three open pretraining datasets (the Pile, the SmolLM corpus, and Refinedweb), we compared the features of their free-generations to features produced by the same taggers and the LM-based classifier (\autoref{appendix:freegen-prompt}) on a randomly sampled 1M document subset of the pretraining corpora. Due to costs, for the domain classification 5k examples from the 1M were used per corpus. The 1M documents were uniformly sampled with reservoir sampling.

\begin{table}[h]
\centering
\small
\label{tab:domain_correlations}
\begin{tabular}{lrr}
\toprule
\textbf{Domain} & \textbf{Pearson r} & \textbf{p-value} \\
\midrule
web & 0.917 & 7.55e-12 \\
reference & 0.832 & 3.99e-08 \\
academic & 0.824 & 7.21e-08 \\
code & 0.679 & 7.14e-05 \\
books & 0.374 & 5.02e-02 \\
\bottomrule
\end{tabular}
\caption{Domain correlations between pretraining data and free-generations. Web content shows the strongest Pearson correlation, suggesting models most faithfully reproduce web distribution patterns, while books content shows the weakest relationship.}

\end{table}

\begin{table}[h]
\centering
\small
\label{tab:feature_correlations}
\begin{tabular}{lrr}
\toprule
\textbf{Feature} & \textbf{Pearson r} & \textbf{p-value} \\
\midrule
conjunctions\_ratio & 0.690 & 4.80e-05 \\
question\_words\_ratio & 0.554 & 2.21e-03 \\
numbers\_ratio & 0.449 & 1.66e-02 \\
imperative\_verbs\_ratio & 0.443 & 1.83e-02 \\
char\_len & 0.214 & 2.75e-01 \\
\bottomrule
\end{tabular}
\caption{Linguistic feature correlations between pretraining data and free-generations. Connective elements like conjunctions show stronger correlations, while structural features like character length are less preserved in model generations.}
\end{table}

Additionally, we calculated two holistic model-wise correlations, which measure how well each model's complete generation profile matches its training data:
\begin{enumerate}
    \item \textbf{Domain-level correlations:} For each domain category (web, code, academic, books, reference), we computed the correlation between the percentage of that domain in the model's documented pretraining data and the percentage of free-generations classified into that category.
    \item \textbf{Feature-level correlations:} For linguistic features (conjunctions ratio, imperative verbs ratio, question words ratio, numbers ratio, character length), we assessed how well each feature in free-generations correlates with corresponding features in pretraining data.
    \end{enumerate}

These model-wise correlations are notably higher than individual feature correlations, indicating that models preserve the overall "fingerprint" of their training data better than any single dimension alone.

\section{Free-generation Examples}
\label{appendix:freegen_examples}

\textbf{Llama-2-70B, web-like:}
Gwen Stefani Couldn’t Recognize Her First Song at an Event with Blake Shelton
What's the first song you ever wrote and or recorded? And how would you feel if you never heard it again until you were Facetiming with hundreds of thousands of people while attending a big country music event?
That's exactly what happened to Gwen Stefani, when she was honored at the Country Radio Seminar event this week. In Nashville, . One can only imagine that when you’ve had as crazy a career as she’s had, it may be a chore to actually remember all of the milestones over the years, but just imagine also doing it on the spot, for everyone to see.
Well, as it turns out, the doubly talented Dolly Parton was set to introduce Gwen as “an artist who broke barriers” when she cited the lyrics from one of Stefani’s first-ever ``country'' songs…``year 3,000.'' The diplomatic ``The Voice'' coach Blake Shelton interjected at the video she was showing at that moment, asking ``was there other work done ....

\textbf{phi-2, books-like:} The sun glistened brightly in the evening sky, casting warm rays across the quaint little kitchen where a pair of diligent parents were preparing a scrumptious meal. The air was filled with anticipation as they embarked on a culinary adventure, their hearts filled with love and joy.

With their sleeves rolled up, the parents deftly wielded an assortment of sharp knives, each one an extension of their skill and precision. Like artists with their brushes, they delicately maneuvered through the ingredients, dancing with knives as their partners. The blades moved with a swiftness that defied the physics of everyday movements, gliding through the air with grace and purpose.

...

\textbf{Qwen2-72B, reference (instruct)-like:}
Children: Family ticket (at most 2 adults and 2 children) \$23.00 (1) Mr Black wants to see the film with his wife and two sons (they're both over 9), how much should he pay?\_\_\_\_ A. \$23. B. \$46. C. \$48. D. \$49. (2)Mr Brown wants to see the film with his 5-year-old twin sons, how much should he pay\_\_\_\_. A. \$18. B. \$23. C. \$28.5 D. \$32 (3)Lisa can't see snow in this film.\_\_\_\_ A. True B. False (4). Jane is a maths teacher and she has two daughters (she's 12 and her sister is 8). They want to see the film. How much should she pay?\_\_\_\_ A. \$28.5 B. \$27 C. \$33 D. \$26.5 (5) Mike is sixteen. Today is Saturday and he wants to see the film with his 40-year-old parents. They will pay\_\_\_\_ for the tickets. A. \$22.

\textbf{OPT-6.7B, web-like:} 
I don't think this is a question. It's a statement. I say get a dog, the world is better with dogs.

I said to myself the exact same thing. I don't know why you were down-voted.

because reddit is a bunch of people who hate dogs.

\newpage
\textbf{Codegen-16B-mono, code-like: }
\begin{lstlisting}[style=pythonstyle]
from typing import Optional

from fastapi import FastAPI
from pydantic import BaseModel

app = FastAPI()


class Item(BaseModel):
    name: str
    price: float
    is_offer: Optional[bool] = None


@app.get("/")
def read_root():
    return {"Hello": "World from FastAPI"}


@app.get("/items/{item_id}")
def read_item(item_id: int, q: Optional[str] = None):
    return {"item_id": item_id, "q": q}


@app.put("/items/{item_id}")
def update_item(item_id: int, item: Item):
    return {"item_name": item.name, "item_id": item_id}
\end{lstlisting}

\textbf{Pythia-410M, academic-like (patent): }
This invention relates to new and improved methods and apparatus for removing solid waste material from the waste stream of a power station where the solid waste material is intended for disposal after completion of the power plant.

Various attempts have previously been made to remove solid waste from the waste stream of plants. This is true, for example, to the discharge of sludge which is generally collected into a sludge tank and washed out of the plant in a washing tank which is generally connected to an open drain outlet of the plant. This prior art is discussed by U.S. Pat. No. 3,623,579 which issued to G. R. Clark and described a method for treating the waste stream to remove solid waste by flocculating and flocculating and agitating the solids in a tank to break bonds between the solid particles.

Furthermore, an apparatus was described by U.S. Pat. No. 4,016,823 which describes a method in which liquid sewage is removed from the waste stream and from the sewage treatment plant where the solid waste being removed is to be treated to produce ammonia-purified water for use in bathing baths or soaps and where the sewage from the wastewater treatment plant is removed to the sewage processing plant where this sewage is mixed with water or treated as a fertilizer.

...

\section{XGBoost Settings}
\label{appendix:xgboost_settings}

For the inner grid search, 
the maximum depth of trees was in $[2, 3, 5]$, while the learning rate was in $[0.01, 0.1, 0.3]$ and the number of trees was in $[50, 100]$.

\section{Selected Features by Task}
\label{appendix:selected_features}

In \autoref{tab:sel_features}, we show the selected features per benchmark.

\begin{table*}[ht]
\centering
\caption{Greedily-selected features per benchmark.}
\label{tab:sel_features}
\begin{tabular}{p{0.25\textwidth} p{0.7\textwidth}}
\toprule
\textbf{Benchmark} & \textbf{Selected Features} \\
\midrule
arc challenge (25-shot) 
& total params, pretraining summary total tokens billions, question words ratio, layer norm type, dimension, pretraining summary percentage code \\
\midrule
gsm8k (5-shot) 
& total params, pretraining summary total tokens billions, pretraining summary percentage reference, edu classifier std, pretraining summary percentage books \\
\midrule
hellaswag (10-shot) 
& total params, pretraining summary total tokens billions, pretraining summary percentage code, pretraining summary percentage reference, positional embeddings, pretraining summary percentage academic \\
\midrule
mmlu 0-shot (0-shot) 
& total params, pretraining summary total tokens billions, layer norm type, activation, pretraining summary percentage code \\
\midrule
truthfulqa (0-shot) 
& total params, pretraining summary total tokens billions, domain web pct mean, dep parse dep root dist max mean, pretraining summary percentage english, entropy mean, layer norm type \\
\midrule
winogrande (5-shot) 
& total params, pretraining summary total tokens billions, question words ratio, layer norm type, pct english mean, positional embeddings, pretraining summary percentage books, pretraining summary percentage code, block type \\
\midrule
anli (0-shot) 
& total params, pretraining summary total tokens billions, pretraining summary percentage code, pretraining summary percentage web, pretraining summary percentage books, positional embeddings \\
\midrule
logiqa2 (0-shot) 
& total params, pretraining summary total tokens billions, pretraining summary percentage web, domain reference pct mean, dep parse dep root dist mean std, dep parse dep root dist median std \\
\midrule
mathqa (5-shot) 
& total params, pretraining summary total tokens billions, pretraining summary percentage books, num heads \\
\midrule
xnli (0-shot) 
& total params, pretraining summary total tokens billions, pretraining summary percentage web \\
\midrule
lambada (0-shot) 
& total params, pretraining summary total tokens billions, pretraining summary percentage code, block type \\
\midrule
mmlu 5-shot (5-shot) 
& total params, pretraining summary total tokens billions, sequence length, biases, num heads, dimension, edu classifier mean, pretraining summary percentage academic \\
\midrule
humaneval (0-shot) 
& total params, pretraining summary total tokens billions, pretraining summary percentage code, layer norm type, pretraining summary percentage english, biases \\
\bottomrule
\end{tabular}
\end{table*}

\section{LightGBM Results}
\label{appendix:lightgbm_ver}

The LightGBM version of \autoref{tab:comparison_final} can be found in \autoref{tab:comparison_lgbm}.

\begin{table*}[!ht]
\centering
\small
\resizebox{0.85\textwidth}{!}{
\begin{tabular}{ll c c c}
\toprule
\textbf{Benchmark} & \textbf{Setting} & \textbf{Baseline MAE} & \textbf{Scaling Laws MAE} & \textbf{All Features MAE} \\
\midrule
\multicolumn{5}{c}{\textbf{Accuracy}} \\
\midrule
Arc Challenge  & 25-shot & 13.23\% & 4.91\% & 3.61\%  \\
GSM8k          & 5-shot  & 15.65\% & 11.03\% & 5.78\%  \\
Hellaswag      & 10-shot & 12.26\% & 4.29\% & 3.14\%  \\
Humaneval      & 0-shot  & 11.79\% & 8.61\% & 6.80\%  \\
Lambada        & 0-shot  & 16.89\% & 9.60\% & 6.39\%  \\
MMLU (0-shot)  & 0-shot  & 11.98\% & 9.12\% & 4.21\%  \\
MMLU (5-shot)  & 5-shot  & 12.25\% & 8.39\% & 3.05\%  \\
TruthfulQA     & 0-shot  &  3.72\% & 3.40\% & 2.61\%  \\
Winogrande     & 5-shot  & 10.14\% & 3.99\% & 3.09\%  \\
\midrule
\multicolumn{5}{c}{\textbf{Brier score (×100)}} \\
\midrule
XNLI           & 0-shot  & 7.22   & 4.70   & 4.32   \\
ANLI           & 0-shot  & 9.48   & 6.14   & 6.05   \\
MathQA         & 0-shot  & 7.57   & 3.89   & 2.95   \\
LogiQA2        & 0-shot  & 12.74  & 8.77   & 8.29   \\
\bottomrule
\end{tabular}
}
\caption{MAE comparison of Scaling Laws and All Features predictors versus Baseline. Note that a significance test was not carried out for LGBM, so this reflects results for one run, though a hyperparameter search is still carried out over the same values as in \autoref{appendix:xgboost_settings} for both predictors. Brier scores are scaled ×100 for comparability. Both predictors here use LGBM.}
\label{tab:comparison_lgbm}
\end{table*}

\section{SHAP Plots for remaining benchmarks}
\label{appendix:all_shap_plots}

\begin{figure}
    \centering
    \includegraphics[width=\linewidth]{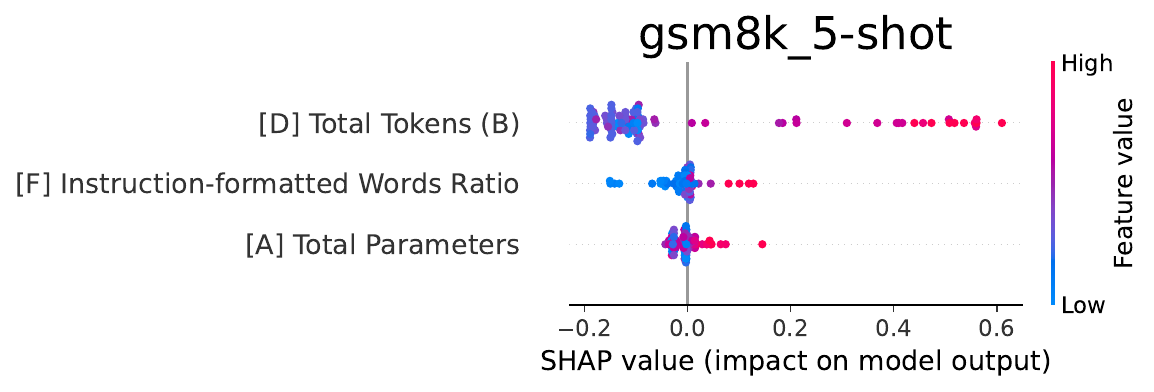}
    \caption{SHAP values for GSM8k}
    \label{fig:shap_1}
\end{figure}

\begin{figure}
    \centering
    \includegraphics[width=\linewidth]{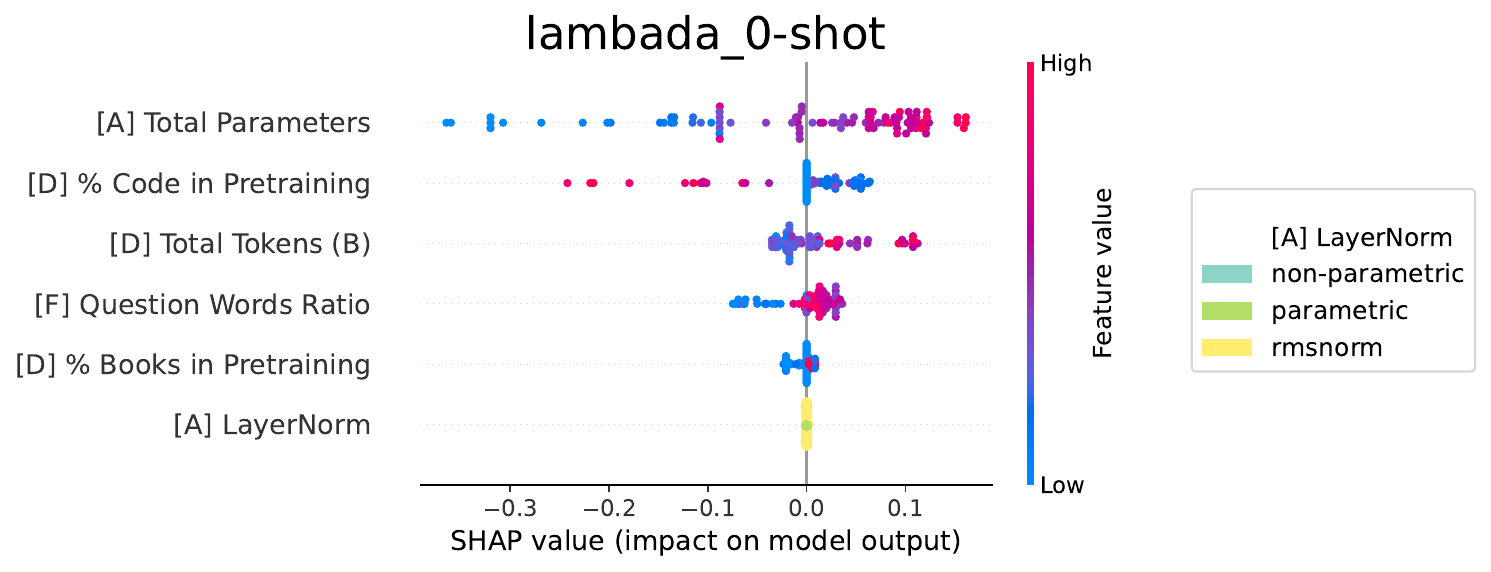}
    \caption{SHAP values for Lambada}
    \label{fig:shap_2}
\end{figure}

\begin{figure}
    \centering
    \includegraphics[width=\linewidth]{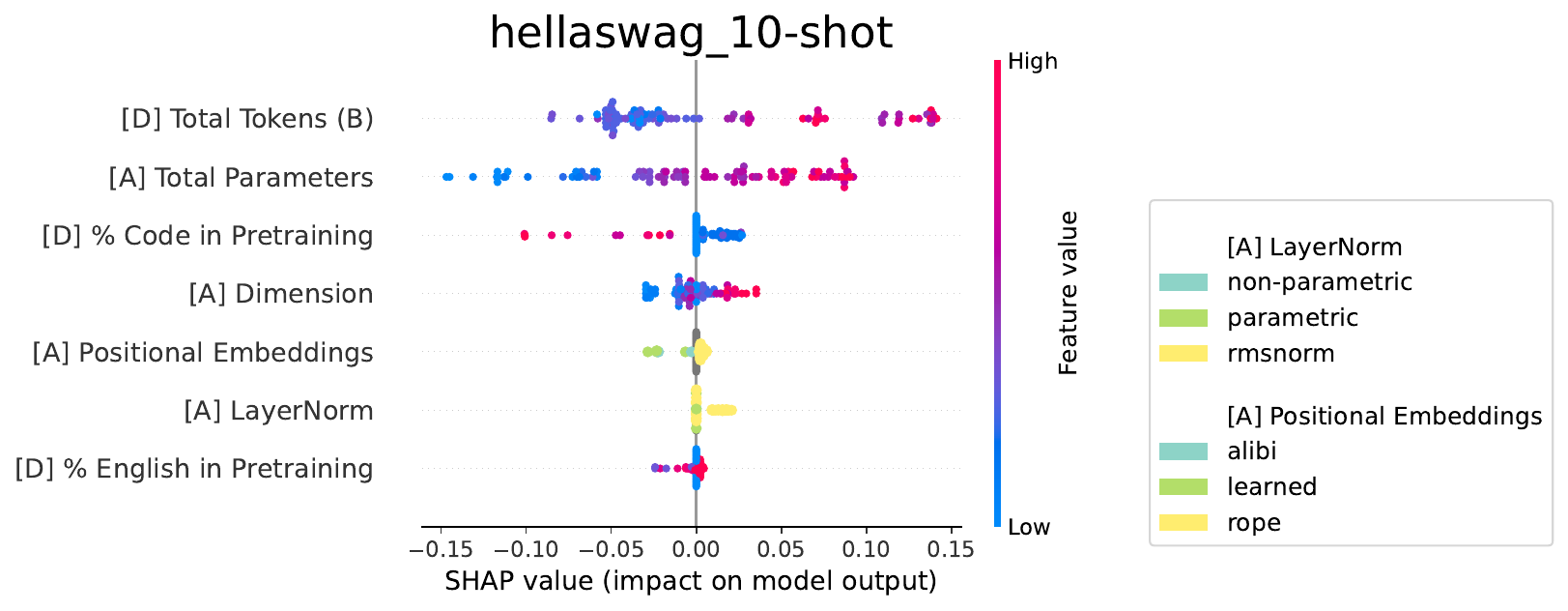}
    \caption{SHAP values for Hellaswag}
    \label{fig:shap_3}
\end{figure}

\begin{figure}
    \centering
    \includegraphics[width=\linewidth]{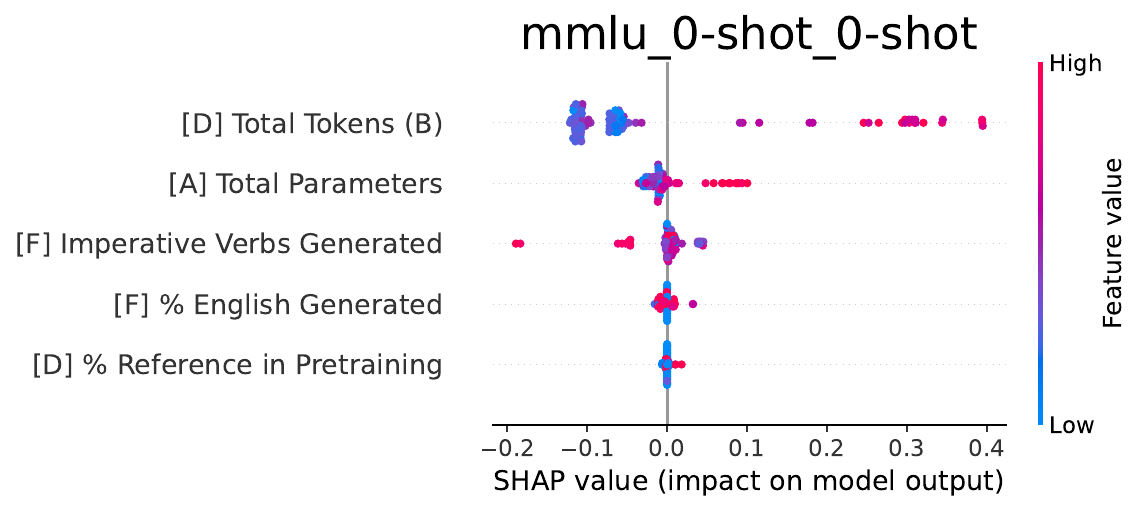}
    \caption{SHAP values for MMLU 0-shot}
    \label{fig:shap_4}
\end{figure}

\begin{figure}
    \centering
    \includegraphics[width=\linewidth]{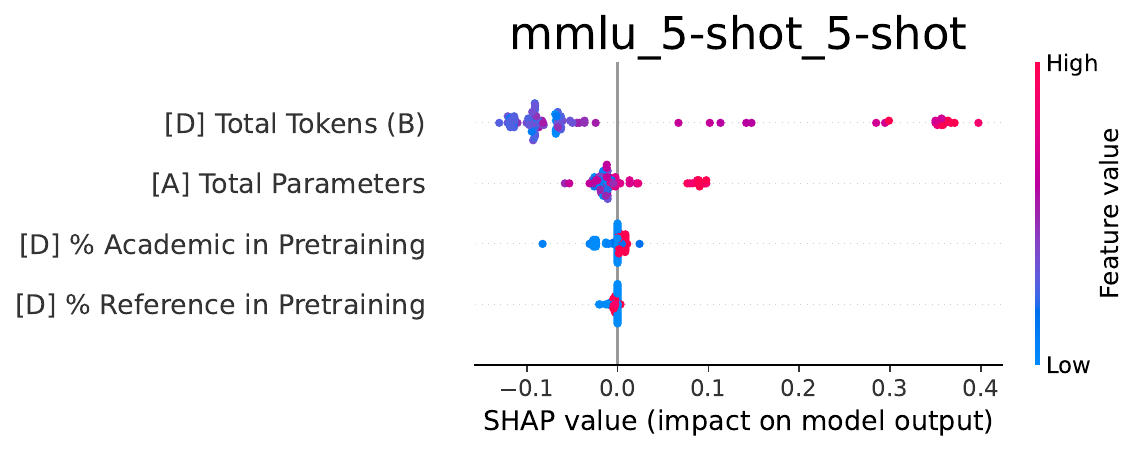}
    \caption{SHAP values for MMLU 5-shot}
    \label{fig:shap_5}
\end{figure}

\begin{figure}
    \centering
    \includegraphics[width=\linewidth]{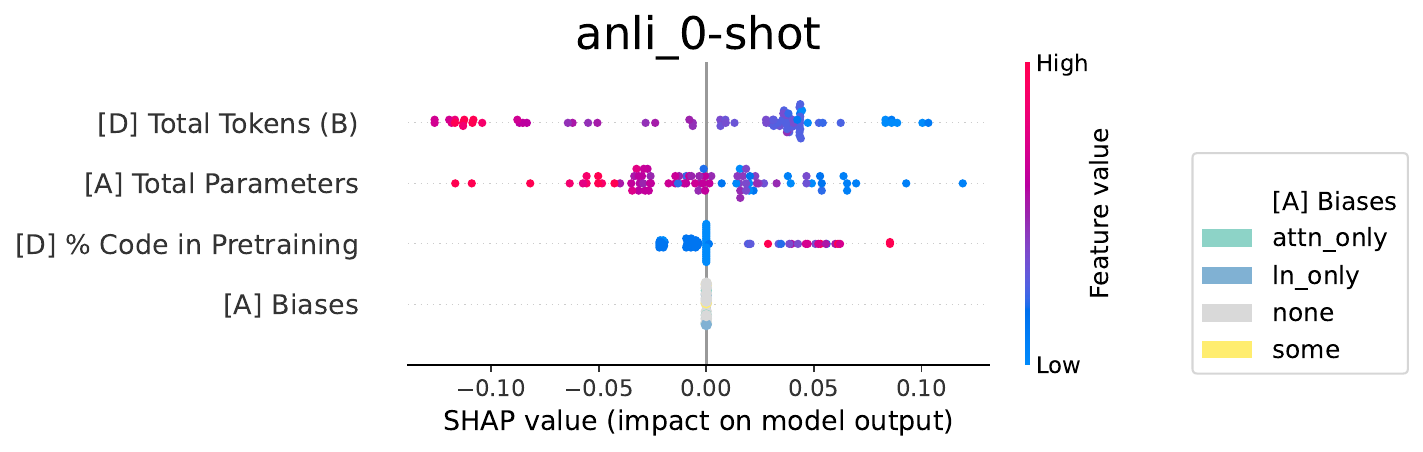}
    \caption{SHAP values for ANLI}
    \label{fig:shap_6}
\end{figure}

\begin{figure}
    \centering
    \includegraphics[width=\linewidth]{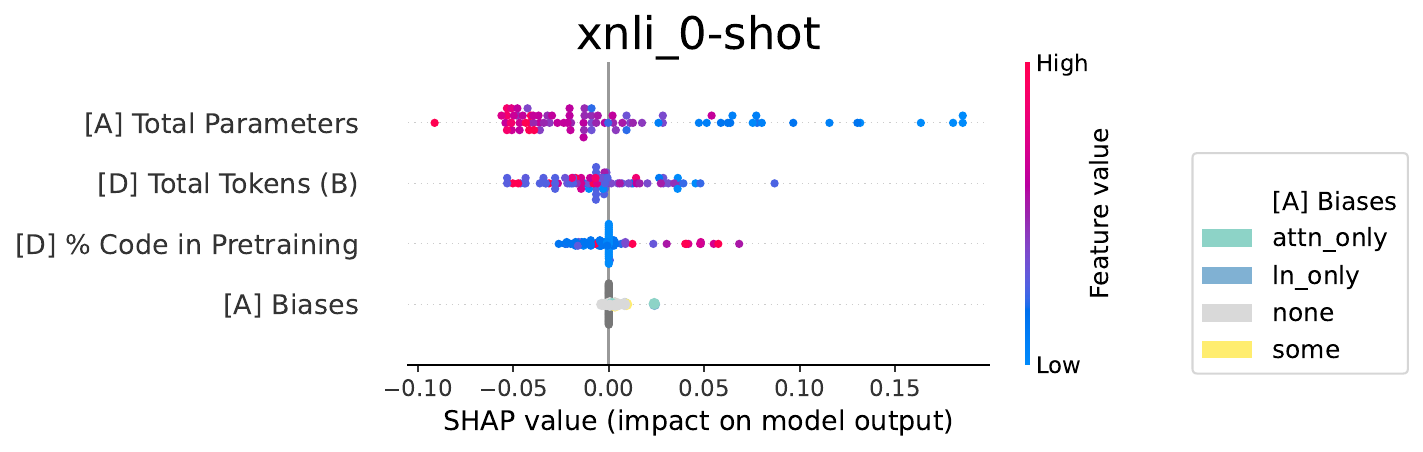}
    \caption{SHAP values for XNLI}
    \label{fig:shap_7}
\end{figure}

\begin{figure}
    \centering
    \includegraphics[width=\linewidth]{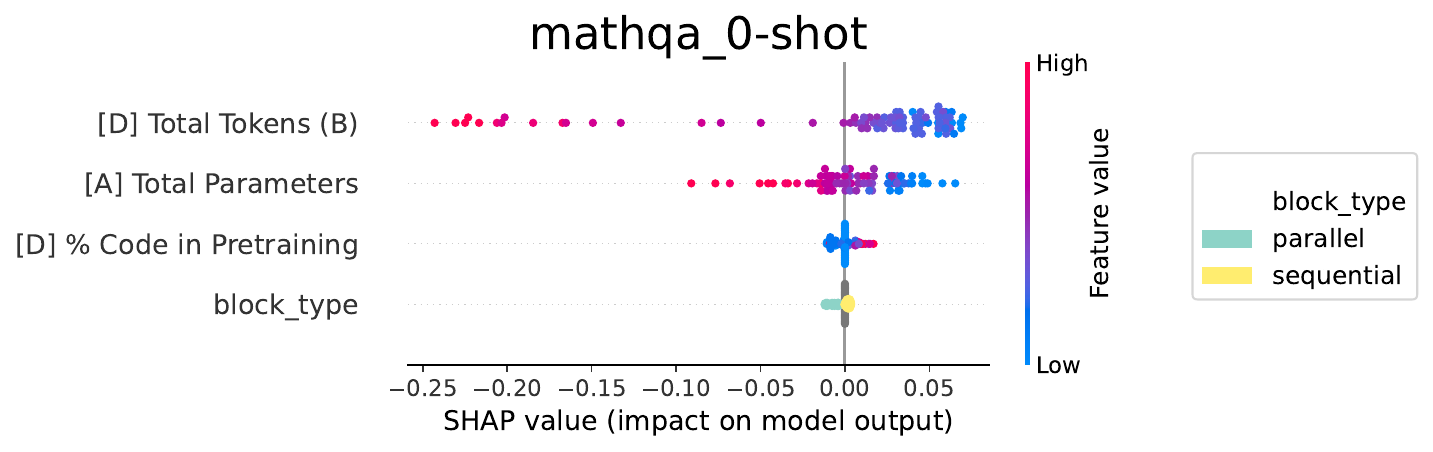}
    \caption{SHAP values for MathQA}
    \label{fig:shap_8}
\end{figure}

\begin{figure}
    \centering
    \includegraphics[width=\linewidth]{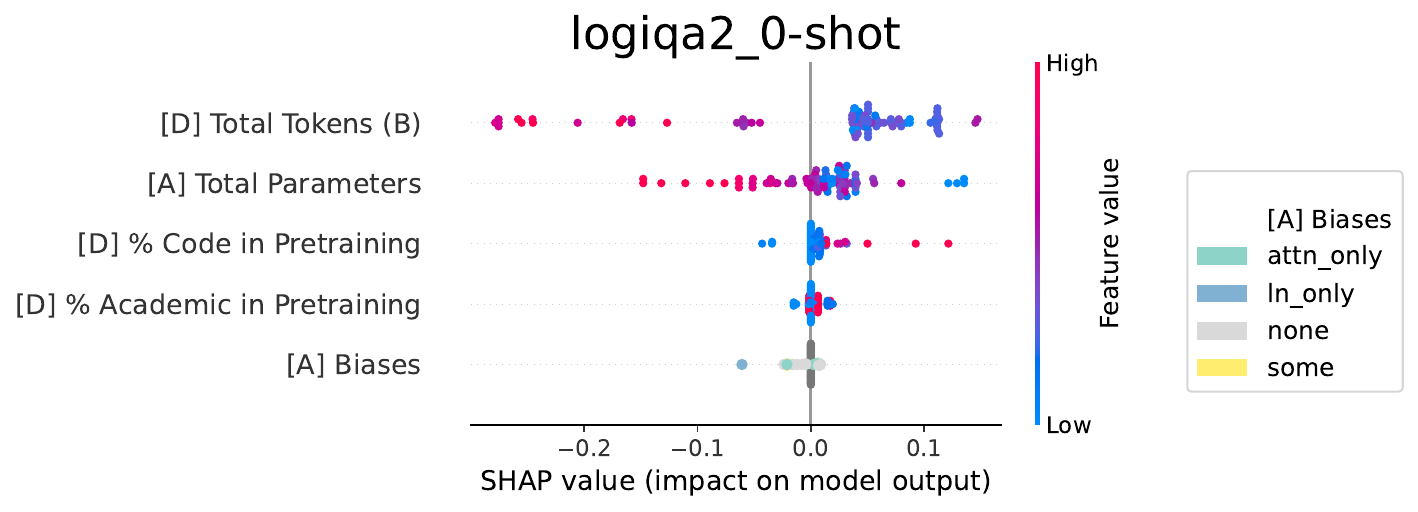}
    \caption{SHAP values for LogiQA2}
    \label{fig:shap_9}
\end{figure}

SHAP plots for the remaining benchmarks can be found in \autoref{fig:shap_1} -- \autoref{fig:shap_9}. Please note that lower scores are better for Brier score tasks (ANLI, XNLI, MathQA, LogiQA2)

\section{Details on confirmatory pretraining runs}
\label{appendix:confirmatory_exps}

\subsection{Training}
For our confirmatory experiments, we trained 460M parameter Llama-2 architecture models from scratch using the Megatron-Deepspeed library. We capped training tokens to 10B, while using a cosine learning rate schedule set to a length of 100B tokens (meaning that each checkpoint is approximately 10\% through a ``full'' pretraining run). Training took place on one node per checkpoint, with 8 H100 GPUs. Each checkpoint took roughly 6 hours to train. 

For our data mixes, we constructed various mixes by using the subsets of the Dolma v1 dataset. In the web vs. other experiments, we fixed the relative percentages of all other data sources while varying the web percentage. 

The training configuration is as follows:
\begin{verbatim}
training:
  num_layers: 14
  num_attention_heads: 12
  seq_length: 2048
  num_kv_heads: 12
  hidden_size: 1536
  ffn_hidden_size: 4128
  tune_steps: 1000
  lr: 0.00015
  min_lr: 1.0e-5
  weight_decay: 1e-2
  grad_clip: 1.0
  lr_warmup_steps: 100
  save_interval: 2000
  eval_interval: 2000
  train_epochs: 1
  tp: 1
  micro_batch_size: 16
  global_batch_size: 512
  seed: 42

\end{verbatim}

Aside from data mix, all experiments used identical hyperparameters to ensure fair comparison. 

\subsection{Evaluation}
To assess the association of different data mixes with model performance, we evaluated our models on the following tasks: 

\begin{enumerate}
    \item \textbf{Natural language inference: } Lambada, winogrande, arc challenge
    \item \textbf{Code generation: } Humaneval
    \item \textbf{Math: } GSM8K
    \item \textbf{Factuality: } TruthfulQA
\end{enumerate}

Note that we do not select the full evaluation set due to time constraints. As LM eval harness does not implement perplexity/loss based evaluations for all tasks, we manually convert multiple-choice tasks to loss-based metrics, and mask out the prompt or question when calculating loss for all tasks.

\subsection{Conversion to Loss-Based Metrics}
To ensure consistent evaluation across different tasks and models, we converted various benchmark datasets to loss-based metrics. This approach allows for more direct comparison between models and clearer interpretation of improvements. Here's how we implemented loss calculations for each dataset type:

\paragraph{Multiple Choice Tasks (ARC Challenge, Winogrande, HellaSwag, TruthfulQA):}
For these datasets, we calculated two primary loss-based metrics:
\begin{itemize}
    \item \textbf{Average Loss:} We computed the negative normalized log probability of the correct answer. For each question, we formatted the input as "Question + Answer Choice", then calculated the sequence log probability normalized by token length for each choice. The negative log probability of the correct answer was used as the loss.
    \item \textbf{Margin-based Loss:} For TruthfulQA specifically, we calculated a margin between truthful and non-truthful answers. This was computed as the negative of the difference between the best truthful answer's log probability and the best non-truthful answer's log probability. A lower loss indicates better differentiation between truthful and non-truthful information.
\end{itemize}

\paragraph{Generation Tasks (GSM8K, HumanEval, Lambada):}
For generation tasks, we calculated:
\begin{itemize}
    \item \textbf{Answer Loss:} We computed the cross-entropy loss on the solution tokens only. Note that for Lambada, this is only the last word.
\end{itemize}

All log probabilities were normalized by sequence length as well.

\subsection{Full Results}

Exact loss values for the code vs. natural language mixes and the web vs. other mixes can be found respectively in \autoref{tab:mix_results_code_nl} and \autoref{tab:mix_results_web_other}.

\begin{table*}[ht]
\centering
\small
\begin{tabular}{lccccc}
\toprule
\textbf{Data Mix} & \textbf{Lambada} & \textbf{Humaneval} & \textbf{Winogrande} & \textbf{Arc challenge} & \textbf{GSM8k} \\
\midrule
NL 70/Code 30 & 3.426 &\textbf{ 1.331} & 3.967 & 3.967 & 2.295 \\
NL 75/Code 25 & 3.426 & \textbf{1.331} & 3.966 & 3.651 & 2.295 \\
NL 80/Code 20 & 3.419 & 1.350 & 3.989 & 3.685 & \textbf{2.293} \\
NL 90/Code 10 & \textbf{3.406} & 1.533 & \textbf{3.944} & \textbf{3.620} & 2.300 \\
\bottomrule
\end{tabular}
\caption{Loss-based evaluation metrics across different data mix ratios}
\label{tab:mix_results_code_nl}
\end{table*}

\begin{table*}[ht]
\centering
\small
\begin{tabular}{lcc}
\toprule
\textbf{Data Mix} & \textbf{Margin Loss} & \textbf{Accuracy (\%)} \\
\midrule
Web 30/Other 70 & 0.2363 & \textbf{28.86} \\
Web 50/Other 50 & 0.\textbf{2342} & 28.40 \\
Web 90/Other 10 & 0.2462 & 28.03 \\
\bottomrule
\end{tabular}
\caption{TruthfulQA evaluation metrics across different web/other ratios}
\label{tab:mix_results_web_other}
\end{table*}

\section{Publication bias correction}
\label{appendix:publication_bias}

Data from PET-PEESE on architectural variations can be found in \autoref{tab:pet-peese}.

\begin{table*}[t]
\centering
\small
\caption{Bias-corrected architecture effects on accuracy (defaults-only; PET-PEESE). Positive values mean the feature level outperforms the rest of the pool within a task. Effects are in percentage points (pp); CIs are 95\%.}
\begin{tabular}{llcccc}
\toprule
Feature & Level & $k$ & Chosen & Effect (pp) [95\% CI] & Egger $p$ \\
\midrule
Positional embeddings & alibi         & 6 & PEESE & $-2.0$ \; [$-7.8$, $+3.9$] & 0.013 \\
Positional embeddings & learned       & 6 & PET   & $+5.4$ \; [$-7.2$, $+17.9$] & 0.240 \\
Positional embeddings & ROPE          & 6 & PEESE & $+1.5$ \; [$+0.9$, $+2.2$] & 0.076 \\
LayerNorm type        & non-parametric& 6 & PEESE & $-3.4$ \; [$-8.6$, $+1.8$] & 0.031 \\
LayerNorm type        & parametric    & 6 & PEESE & $-1.3$ \; [$-2.7$, $-0.0$] & 0.021 \\
LayerNorm type        & RMSNorm       & 6 & PET   & $+2.3$ \; [$+0.9$, $+3.8$] & 0.133 \\
Attention variant     & full          & 6 & PEESE & $+1.1$ \; [$+0.2$, $+2.0$] & 0.014 \\
Attention variant     & GQA           & 6 & PET   & $+3.4$ \; [$+0.6$, $+6.3$] & 0.290 \\
Attention variant     & local,full    & 6 & PET   & $+1.3$ \; [$-1.8$, $+4.3$] & 0.358 \\
Attention variant     & MQA           & 6 & PEESE & $0.0$ \; [$0.0$, $0.0$]$^\dagger$ & 0.038 \\
\bottomrule
\end{tabular}
\label{tab:pet-peese}
\end{table*}

\end{document}